\PassOptionsToPackage{table,dvipsnames}{xcolor}
\documentclass[11pt]{article}
\usepackage[round,authoryear]{natbib}
\usepackage{metacircle}

\usepackage{amsmath,amsfonts,bm}

\def\eqref#1{equation~\ref{#1}}

\def\1{\bm{1}}

\DeclareMathAlphabet{\mathsfit}{\encodingdefault}{\sfdefault}{m}{sl}
\SetMathAlphabet{\mathsfit}{bold}{\encodingdefault}{\sfdefault}{bx}{n}

\usepackage{placeins}
\usepackage{tabularx}
\usepackage{float}
\usepackage{multirow}

\title{Meta Models: Models that Model Learning}
\author{%
  Li Peihang\textsuperscript{1,2,3}\quad
  Liu Ziming\textsuperscript{1,4,5,*}\\
  {\small\textsuperscript{1}MetaCircle\quad
  \textsuperscript{2}The University of Hong Kong\quad
  \textsuperscript{3}University of California, Berkeley}\\
  {\small\textsuperscript{4}Shanghai Qizhi Institute\quad
  \textsuperscript{5}Tsinghua University}\\
  {\small\textsuperscript{*}Corresponding author}%
}

\date{}
\paperdate{October 12, 2026}
\papertype{Meta Circle Research}
\projecturl{}
\hypersetup{pdftitle={Meta Models: Models that Model Learning},
  pdfauthor={Li Peihang and Liu Ziming}}

\begin{document}

\maketitle

\begin{metaabstract}
Every neural-network training run produces not only a model, but also evidence about how models learn. Can these accumulated experiments enable us to predict and compare new training runs before executing them? We study meta-models: models that predict the learning behavior of other models. From this perspective, final-performance prediction, learning-curve prediction, and hyperparameter selection are different queries to the same training response function. We take a first, humble empirical step toward this goal: treating real language-model training records as native numeric data, we find that an off-the-shelf tabular in-context learner, TabPFN, can match or outperform a fine-tuned LLM meta-model in multiple evaluated settings. Whereas recent work fine-tunes LLMs on tokenized configurations at a cost of tens to hundreds of GPU-hours, TabPFN requires seconds of fit-and-predict time without task-specific weight updates. We argue that meta-modeling is better treated as its own numeric modality rather than as a text task for LLMs. Going one step further, augmenting TabPFN with Kolmogorov--Arnold Networks (KANs) further improves accuracy on selected prediction tasks. Finally, we pre-registered a forecast of the Marin Hero Run, a 535B-parameter public training run, whose ground truth is revealed point by point as training progresses, putting meta-model predictions to the test against training outcomes not yet observed at the time of forecasting.
\end{metaabstract}

\FloatBarrier
\section{Introduction}
\label{sec:intro}

\begin{figure}[t]
\centering
\includegraphics[width=\linewidth]{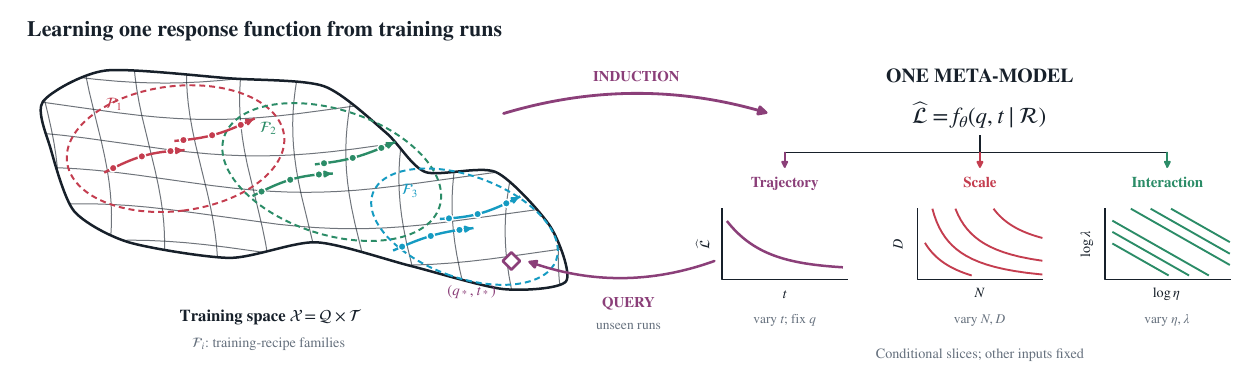}
\caption{\textbf{Meta-modeling as induction over the space of training runs.} Paths represent runs, dots their observations, and the diamond an unseen query in configuration--progress space. Dashed regions denote training-recipe families. A meta-model predicts loss from historical records; trajectory, scale, and interaction views are conditional slices of this same function. Geometry and slices are schematic, not measured embeddings or results.}
\label{fig:overview}
\end{figure}

Before launching a training run, researchers already make predictions. Will a different learning rate help? Which data recipe should receive the next compute budget? Will a promising small-scale result survive scaling up? Each completed run supplies evidence for such judgments. Can these accumulated records become a reusable model of training, allowing new experiments to be queried before they are executed?

We call this predictive object a \emph{meta-model of training}: a model that predicts the learning behavior of other models. In our loss-prediction setting,
\[
\widehat{\mathcal{L}}(q,t)=f_\theta(q,t\mid\mathcal{R}),
\qquad
\mathcal{R}=\{(q_i,t_{ij},\mathcal{L}_{ij})\}_{i,j},
\]
where $q$ is a training configuration, $t$ is progress, and $\mathcal{R}$ contains historical observations. Querying different progress values produces a trajectory; comparing configurations supports experiment selection (Figure~\ref{fig:overview}). Other measured outcomes, such as downstream accuracy, define further prediction tasks.

Scaling laws connect loss to model and data size \citep{kaplan2020scaling,hoffmann2022chinchilla}; normalized curves reveal shared dynamics under compute-optimal training \citep{qiu2025collapse}. NCPL extends this line of work by using a fine-tuned LLM to represent a configuration-to-performance scaling law \citep{ncpl2026}. These advances motivate a central challenge for meta-modeling: combining accurate prediction, broad applicability, and efficient adaptation. A useful meta-model should accurately predict the outcomes of training choices, support diverse configurations and prediction targets, and incorporate new observations without repeated task-specific retraining.

We take a first, humble empirical step by treating language-model training records as a \emph{native numeric modality}: configuration fields and measurements form input--output examples rather than tokenized descriptions. We find that TabPFN, a pretrained tabular model that learns predictive priors from synthetic tables \citep{hollmann2025tabpfn,grinsztajn2026tabpfn3}, already supports useful predictions of training. Across tasks, we supply historical examples as context and new configurations as queries, reusing the same pretrained backbone without updating its weights.

Our experiments reveal three capabilities of this meta-model (\S\ref{sec:shootout}): \textbf{accurate prediction of training behavior}, encompassing final loss and configuration-dependent trajectories, with further applications to other training observables and downstream accuracy; \textbf{reuse across training questions}, as the same pretrained backbone extends from dense-model prediction to MoE expert configurations, data recipes, and different prediction targets without task-specific fine-tuning; and \textbf{adaptation through new observations}, where adding the first $20\%$ of each target trajectory to historical context reduces run-averaged future-loss MAE by $56\%$ across seven held-out Delphi runs \citep{held2026delphi}, without updating backbone weights.

We further use KANs \citep{liu2024kan} to learn inspectable configuration coordinates that improve StepLaw \citep{li2025steplaw} final-loss and curve prediction without updating TabPFN's backbone (\S\ref{sec:better}). To move beyond retrospective evaluation, we also froze a forecast during the Marin Hero Run and evaluated it against subsequently arriving observations (\S\ref{sec:marin_hero}). These studies also expose a remaining challenge: reliable extrapolation beyond observed training conditions.

\FloatBarrier

\section{TabPFN is already a strong meta-model}
\label{sec:shootout}

We test whether one frozen TabPFN backbone distinguishes training choices, forecasts trajectories, predicts downstream performance, and selects configurations to evaluate. Configuration fields, targets, and historical records change between tasks. Each supplies its own labeled context: the claim is reuse of a numerical prior across training questions, not prediction without relevant records.

Final-loss prediction provides a common reference. Table~\ref{tab:sec2_summary} selects each method by ID MAE separately for final loss and curves, restricting TabPFN to zero-shot variants. TabPFN has lower final-loss MAE than both paper-reported and reproduced NCPL on Marin and StepLaw ID. On OOD, it outperforms reproduced NCPL on Marin but not StepLaw; paper-reported NCPL has lower MAE on both. The same frozen backbone also predicts outcomes for expert configurations and data recipes. We distinguish the released NCPL interface from an extension explicitly fine-tuned for MoE; its limitations are not impossibility results for LLMs.

We retain the run-level splits and released NCPL checkpoints of \citet{ncpl2026}. Marin \citep{wen2025optimizers} and StepLaw \citep{li2025steplaw} vary scale and optimizer settings; Marin-MoE adds expert configurations, and DataDecide \citep{magnusson2025datadecide} varies data recipes. Contexts and target processing are task-specific. TabPFN predicts loss residuals, added back to corpus-specific scaling-law offsets. No queried-run observations are supplied unless stated. ID is validation within observed model sizes, used for selection; OOD contains larger models. Loss definitions remain corpus-specific, so errors are compared within corpora (Appendix~\ref{app:sec2_inventory}).

\begin{table}[!t]
\centering
\caption{\textbf{Predicting final losses and training trajectories.} One entry per method, selected by ID MAE separately for final loss and curves; all corresponding ID/OOD metrics are retained. MAE and RMSE: lower is better; $\rho$: higher is better. Bold marks the best displayed value within each corpus, including paper-reported results; it does not indicate significance.}
\label{tab:sec2_summary}
\begingroup
\scriptsize
\setlength{\tabcolsep}{2.3pt}
\renewcommand{\arraystretch}{1.04}
\begin{tabularx}{\linewidth}{@{}l*{10}{>{\raggedleft\arraybackslash}X}@{}}
\toprule
 & \multicolumn{3}{c}{Final loss (ID)} & \multicolumn{3}{c}{Final loss (OOD)} & \multicolumn{2}{c}{Curve (ID)} & \multicolumn{2}{c}{Curve (OOD)} \\
\cmidrule(lr){2-4}\cmidrule(lr){5-7}\cmidrule(lr){8-9}\cmidrule(l){10-11}
Method & MAE $\downarrow$ & RMSE $\downarrow$ & $\rho\uparrow$ & MAE $\downarrow$ & RMSE $\downarrow$ & $\rho\uparrow$ & MAE $\downarrow$ & RMSE $\downarrow$ & MAE $\downarrow$ & RMSE $\downarrow$ \\
\midrule
\multicolumn{11}{@{}l}{\textbf{Marin}} \\
NCPL (paper) & 0.0109 & 0.0169 & 0.9813 & \textbf{0.0168} & \textbf{0.0239} & 0.9299 & 0.0252 & \textbf{0.0475} & 0.0363 & 0.0644 \\
NCPL (repro) & 0.0115 & 0.0182 & 0.9789 & 0.0189 & 0.0290 & 0.9474 & 0.0312 & 0.0516 & \textbf{0.0350} & \textbf{0.0642} \\
XGBoost & 0.0147 & 0.0239 & 0.9621 & 0.0356 & 0.0408 & 0.9063 & \textbf{0.0242} & 0.0479 & 0.0423 & 0.0655 \\
Parametric law & 0.0285 & 0.0392 & 0.8986 & 0.0205 & 0.0276 & 0.7670 & 0.8036 & 2.3168 & 0.5821 & 1.8930 \\
\rowcolor[HTML]{EAF4F6}\textbf{TabPFN} & \textbf{0.0080} & \textbf{0.0140} & \textbf{0.9872} & 0.0178 & 0.0249 & \textbf{0.9580} & 0.0283 & 0.0588 & 0.0457 & 0.1360 \\
\midrule
\multicolumn{11}{@{}l}{\textbf{StepLaw}} \\
NCPL (paper) & 0.0097 & 0.0158 & 0.9948 & 0.0223 & 0.0345 & 0.9837 & 0.0278 & 0.0981 & 0.0415 & 0.1696 \\
NCPL (repro) & 0.0096 & 0.0152 & 0.9957 & \textbf{0.0204} & \textbf{0.0287} & \textbf{0.9893} & 0.0299 & 0.0972 & 0.0392 & \textbf{0.1690} \\
XGBoost & 0.0103 & 0.0160 & 0.9954 & 0.0226 & 0.0313 & 0.9860 & 0.0293 & 0.1025 & 0.0390 & 0.1741 \\
Parametric law & 0.0234 & 0.0390 & 0.9768 & 0.0257 & 0.0377 & 0.9818 & 0.1612 & 0.2682 & 0.1399 & 0.2795 \\
\rowcolor[HTML]{EAF4F6}\textbf{TabPFN} & \textbf{0.0063} & \textbf{0.0102} & \textbf{0.9981} & 0.0227 & 0.0311 & 0.9852 & \textbf{0.0165} & \textbf{0.0959} & \textbf{0.0367} & 0.1740 \\
\midrule
\multicolumn{11}{@{}l}{\textbf{Marin-MoE}} \\
NCPL (MoE FT)$^{*}$ & 0.0338 & 0.0529 & 0.9597 & 0.0595 & 0.0649 & 0.8495 & 0.0446 & 0.1269 & 0.1086 & 0.1261 \\
Parametric law & 0.0574 & 0.0722 & 0.9515 & 0.0579 & 0.0786 & 0.7386 & 0.5688 & 1.5075 & 0.2055 & 0.3478 \\
Sparsity-aware law & 0.0365 & 0.0615 & \textbf{0.9676} & 0.0350 & 0.0453 & 0.7635 & 0.5683 & 1.5071 & 0.2502 & 0.3828 \\
\rowcolor[HTML]{EAF4F6}\textbf{TabPFN} & \textbf{0.0160} & \textbf{0.0278} & 0.9627 & \textbf{0.0187} & \textbf{0.0272} & \textbf{0.9233} & \textbf{0.0259} & \textbf{0.0614} & \textbf{0.0326} & \textbf{0.0469} \\
\midrule
\multicolumn{11}{@{}l}{\textbf{DataDecide}} \\
NCPL (transfer)$^{*}$ & 0.1023 & 0.1288 & 0.9650 & 0.1247 & 0.1517 & 0.3360 & 0.0994 & 0.1257 & 0.1364 & 0.1661 \\
XGBoost & 0.0136 & 0.0200 & 0.9965 & 0.0202 & 0.0298 & \textbf{0.9916} & 0.0155 & 0.0234 & 0.0232 & 0.0344 \\
Parametric law & 0.0797 & 0.0960 & 0.9650 & 0.1102 & 0.1541 & 0.3360 & 0.1108 & 0.1423 & 0.1389 & 0.1770 \\
\rowcolor[HTML]{EAF4F6}\textbf{TabPFN} & \textbf{0.0105} & \textbf{0.0160} & \textbf{0.9972} & \textbf{0.0178} & \textbf{0.0213} & 0.9899 & \textbf{0.0125} & \textbf{0.0224} & \textbf{0.0133} & \textbf{0.0157} \\
\bottomrule
\end{tabularx}
\endgroup
\par\smallskip
\begin{minipage}{\linewidth}\footnotesize
NCPL (paper): values reported by \citet{ncpl2026}; NCPL (repro): our evaluation of released checkpoints. $^{*}$MoE uses our one-seed fine-tune; DataDecide uses the released model without adaptation. Neither has a paper-reported counterpart. TabPFN uses zero-shot variants only: frozen pretrained weights with labeled training runs as context, excluding the doc-guided tuning ladder. Settings and selection details: Appendix~\ref{app:t1notes}.
\end{minipage}
\end{table}

A separate joint-context benchmark answers 1,905 final-loss queries in $1.79$ seconds on one H200; this is not an end-to-end timing of Table~\ref{tab:sec2_summary}'s selection and corpus-specific prediction routes. Cost accounting is in Table~\ref{tab:cost}; matched-data comparisons are in Appendix~\ref{app:sec2_dataeff}.

\subsection{Predicting differences between training configurations}
\label{sec:sec2_prediction}

At a fixed model size and training budget, can a meta-model distinguish better training configurations from worse ones? We test this capability on optimizer settings and data mixtures, withholding the outcomes of the queried configurations.

\enlargethispage{\baselineskip}

\begin{figure}[!htbp]
\centering
\includegraphics[width=\linewidth]{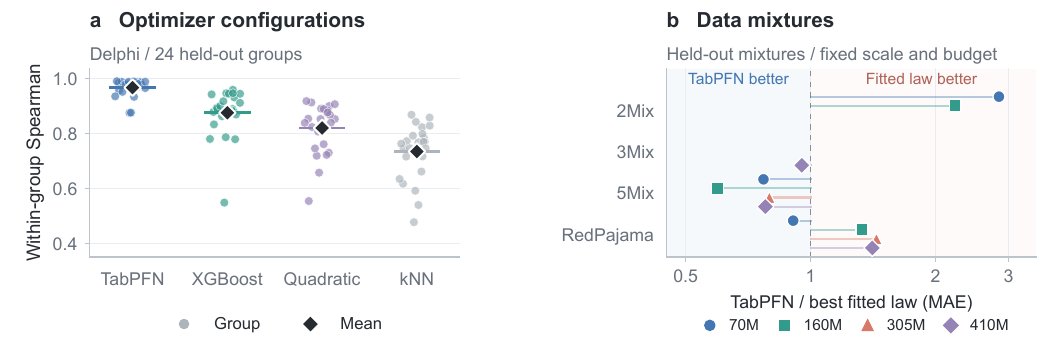}
\caption{\textbf{Predicting configuration differences at fixed scale and budget.} (a) Delphi: each dot is the within-group rank correlation for one of all 24 held-out training-condition groups; diamonds are means. Losses are clipped at $6.0$. (b) Leave-one-mixture-out MAE relative to the lowest evaluated MAE among fitted M1--M4 laws in each family--size cell. All eleven cells appear; markers denote model sizes. Below $1$ favors TabPFN. Ratios are point estimates, not significance tests. Aggregation and fitting protocols: Appendices~\ref{app:sec2_capability_protocol} and~\ref{app:mixlaws}.}
\label{fig:sec2_configurations}
\end{figure}

Delphi's $3{,}737$-run hyperparameter sweep \citep{held2026delphi}, denoted Delphi-HP, lets us test whether this prediction transfers between training conditions. Inputs are model width, batch size, token budget, and seven optimizer hyperparameters; the output is final Paloma \citep{magnusson2023paloma} macro evaluation loss. We withhold all outcomes at one width--batch--budget combination and give TabPFN only the configuration--loss records from the other 23 combinations. Within each held-out group, we rank configurations by predicted loss. Across all 24 folds, mean Spearman correlation is $0.967$, versus $0.875$ for XGBoost \citep{chen2016xgboost} (Figure~\ref{fig:sec2_configurations}a); TabPFN also has lower loss MAE in every group.

The second task predicts how training-data proportions affect evaluation loss. At fixed scale and progress, we withhold one mixture's losses and predict them from its proportions and other mixtures' records. Targets include domain losses and, where defined, total loss. On mixtures of five training domains from \citet{ye2025mixinglaws}, TabPFN's MAE is approximately $20$--$40\%$ below the best evaluated parametric law at each of four model sizes (Figure~\ref{fig:sec2_configurations}b). The laws have lower error on two-domain mixtures and three of four RedPajama sizes; the three-domain comparison is close. The frozen backbone thus predicts differences between training choices, not only between larger and smaller runs.

\subsection{Predicting the course of training}
\label{sec:sec2_dynamics}

Can a meta-model forecast how a training run will unfold before observing its losses? We provide historical runs as context and query TabPFN with a configuration and training progress; successive queries reconstruct a loss trajectory. All evaluations withhold the queried run's losses and reuse the same frozen backbone, with corpus-specific features and context.

\begin{figure}[!t]
\centering
\includegraphics[width=\linewidth]{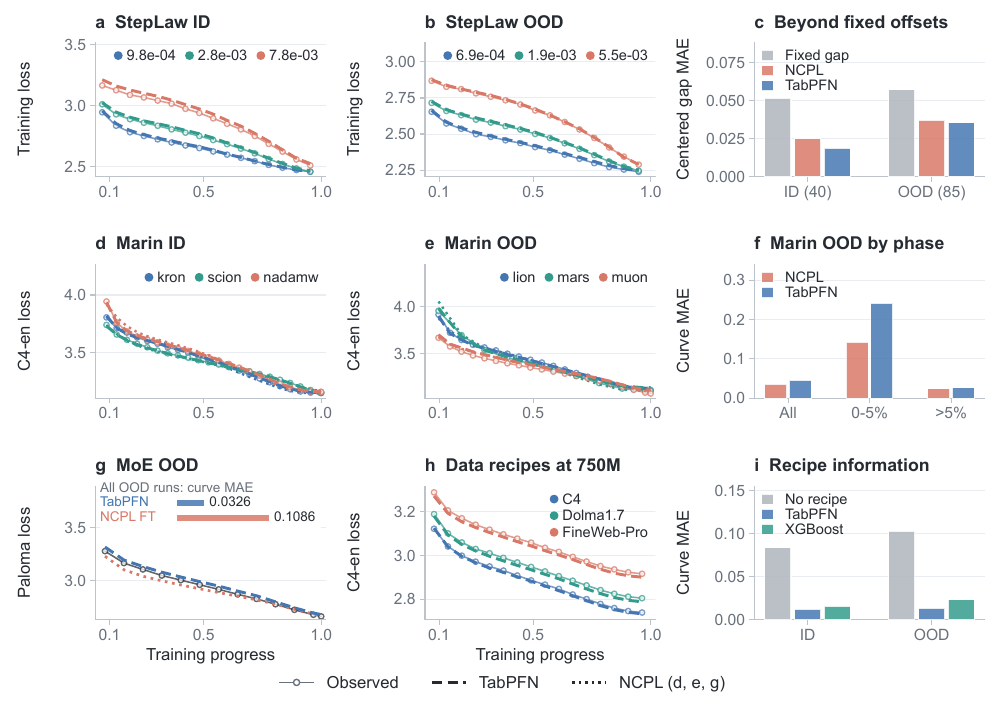}
\caption{\textbf{One frozen backbone forecasts configuration-dependent training trajectories.} (a--b) StepLaw learning-rate sweeps; (c) trajectory differences beyond fixed offsets. (d--e) Marin optimizer configurations; (f) an early-phase OOD weakness. (g) MoE trajectory and aggregate OOD MAE, compared with MoE-fine-tuned NCPL. (h) Data recipes at a held-out 750M scale; (i) recipe ablation and XGBoost comparison. Colors distinguish configurations in a--b, d--e, h; line styles distinguish predictors. Metadata-selected curves display progress $>5\%$; absolute-loss aggregates retain all checkpoints. No target-run losses are supplied. Point estimates; protocols and expanded plots: Appendix~\ref{app:sec2_capability_protocol}.}
\label{fig:sec2_dynamics}
\end{figure}

\textbf{Configuration-dependent dynamics.} TabPFN captures how learning-rate-dependent loss gaps narrow during StepLaw training (Figure~\ref{fig:sec2_dynamics}a--b). These forecasts go beyond shifting a common decay curve by a fixed offset: their error on time-varying OOD gaps is $38\%$ lower than this control (c). Marin extends the test to different optimizers (d--e). Accuracy is not uniform, however: TabPFN's higher Marin OOD error than reproduced NCPL arises mainly in early training (f; Table~\ref{tab:sec2_summary}).

\textbf{Extending to expert configurations.} Adding expert count and routing settings lets the same backbone forecast Marin-MoE trajectories from historical MoE runs (Figure~\ref{fig:sec2_dynamics}g). Across held-out OOD runs, curve MAE is approximately $70\%$ lower than MoE-fine-tuned NCPL; this advantage persists beyond early training. The fine-tuned baseline lacks expert fields, so this comparison evaluates the systems as used, rather than isolating backbone quality under identical inputs.

\textbf{Transferring data recipes across scales.} On DataDecide, records from smaller models support trajectory forecasts for held-out 750M--1B models. TabPFN distinguishes the direction and approximate magnitude of recipe differences at a common scale (Figure~\ref{fig:sec2_dynamics}h), with lower error than recipe-aware XGBoost (i). Removing recipe identity increases OOD error nearly eightfold: the forecasts use recipe information rather than scale alone. All evaluated recipes occur in context; the transfer is to larger models, not unseen recipes.

\subsection{Predicting outcomes beyond loss}
\label{sec:sec2_observables}

\begin{figure}[!htbp]
\centering
\includegraphics[width=\linewidth]{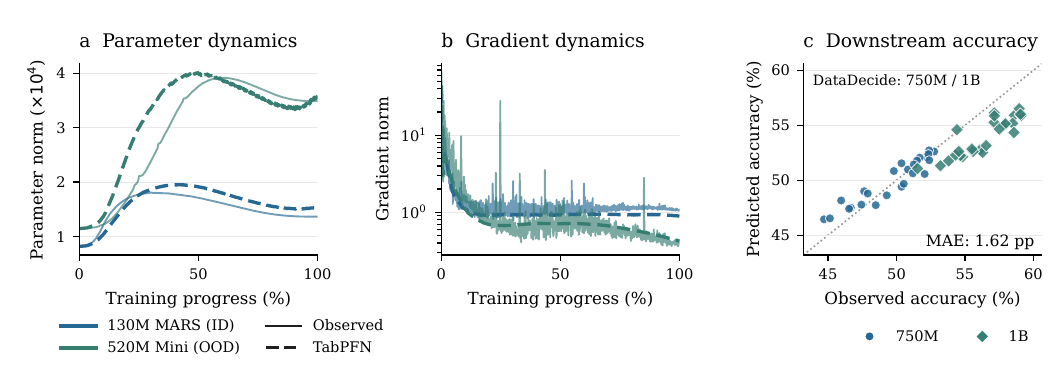}
\caption{\textbf{Predicting internal dynamics and downstream outcomes.} (a--b) Recorded (solid) and predicted (dashed) norms for the same ID and OOD runs; norms sum per-tensor L2 norms, with a log axis in (b). (c) Each point is a held-out DataDecide size--recipe group; the dotted line is perfect prediction. Targets are the last available benchmark evaluations, not necessarily completed training. All panels average two TabPFN seeds. Selection rules and scoring: Appendix~\ref{app:sec2_capability_protocol}.}
\label{fig:sec2_downstream}
\end{figure}

\textbf{Internal training observables.} Can a meta-model predict how training evolves beyond the loss curve? We use Marin records to predict parameter and gradient norms from configuration and training progress, without observing the queried run's weights or trajectory. Figure~\ref{fig:sec2_downstream}a--b compares predicted and recorded trajectories for the same two held-out runs. TabPFN captures the rise and subsequent decline of parameter norms and the early decay of gradient norms, rather than imposing a shared decreasing curve on both targets. Peak timing and magnitude remain imperfect, and the forecasts do not recover individual gradient spikes. These predictions reuse the frozen backbone with observable-specific contexts and target processing.

\textbf{Downstream performance.} We next ask whether configuration--accuracy records from smaller models can forecast the benchmark scores of larger models. We supply historical checkpoint evaluations as context and query DataDecide's held-out 750M--1B models, without observing their losses or benchmark scores. Figure~\ref{fig:sec2_downstream}c shows that TabPFN distinguishes downstream performance across configurations, although it still underestimates the strongest 1B outcomes. Ten-task OLMES \citep{gu2024olmes} macro-accuracy MAE is $1.62$ percentage points, versus $3.40$ for XGBoost using the same historical records. Errors are lower on all ten individual tasks than under the earlier endpoint-context protocol. Together, these results extend the meta-model's queries from loss to internal dynamics and downstream outcomes without training a new neural backbone for each target.

\subsection{Adapting predictions and guiding experiments}
\label{sec:sec2_decisions}

\begin{figure}[!htbp]
\centering
\includegraphics[width=\linewidth]{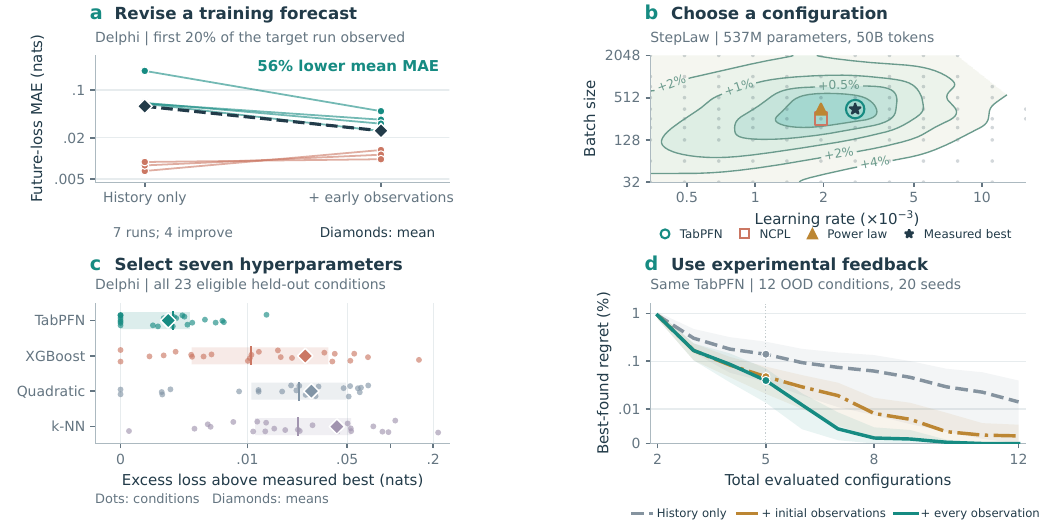}
\caption{\textbf{Revise forecasts and guide training decisions.} (a) All seven Delphi runs; diamonds mark means. (b) A StepLaw sweep; contours interpolate measured loss. (c) All 23 Delphi conditions; diamonds mark means, boxes the interquartile range. (d) Same-backbone context controls; 95\% bootstrap bands over twelve conditions. Budgets include initialization; regret uses measured minima. Log axis in (a), symmetric-log regret in (c--d). Protocols: Appendix~\ref{app:sec2_capability_protocol}.}
\label{fig:sec2_selection}
\end{figure}

\textbf{Updating forecasts during training.} We supply historical training records as context and query losses at future steps for a target configuration. As target-run measurements arrive, we append these configuration--step--loss records and query the remaining trajectory again, without updating backbone weights. On Delphi's seven larger target runs, observing the first $20\%$ of training reduces average remaining-curve error by $56\%$ (Figure~\ref{fig:sec2_selection}a). The benefit is not uniform across runs, but demonstrates that forecasts can be revised through observations rather than another round of model training.

\textbf{Choosing training configurations.} Before launching a run, we query candidate configurations and select the one with the lowest predicted final loss. Historical configuration--loss pairs supply the context; candidate losses are hidden during selection and used afterward to score the recommendation against the best measured candidate. StepLaw's learning-rate and batch-size sweeps illustrate this procedure (Figure~\ref{fig:sec2_selection}b). It also applies to Delphi's seven-dimensional optimizer search: across 23 held-out condition groups, mean excess loss is $0.0038$ nats, versus $0.0254$ for XGBoost (c). On StepLaw, TabPFN can select even better: with the learned configuration coordinates of Section~\ref{sec:better}, its selection regret on larger-model sweeps falls by $80\%$.

\textbf{Using observations to guide subsequent experiments.} In offline StepLaw replay, we reveal each selected candidate's measured loss, append it to context, and use updated predictions to choose the next untested configuration. All policies start from the same two random evaluations. We compare history-only predictions, conditioning on the two initial outcomes but freezing context thereafter, and continual context updates. After five total evaluations, mean OOD best-found regret is $0.1398\%$, $0.0473\%$, and $0.0394\%$, respectively (Figure~\ref{fig:sec2_selection}d). Incorporating target-condition observations brings the search closer to the best measured configuration with fewer evaluations than history-only predictions, without retraining the backbone.

We next ask whether learned configuration coordinates can improve this predictor.

\FloatBarrier

\section{TabPFN could be better}
\label{sec:better}

Good hyperparameters change with training scale. We use Kolmogorov--Arnold Networks (KANs; \citealp{liu2024kan}) to learn scale-dependent learning-rate and batch-size references, making a configuration's position relative to favorable choices explicit.

\begin{figure}[!htbp]
\centering
\includegraphics[width=\linewidth]{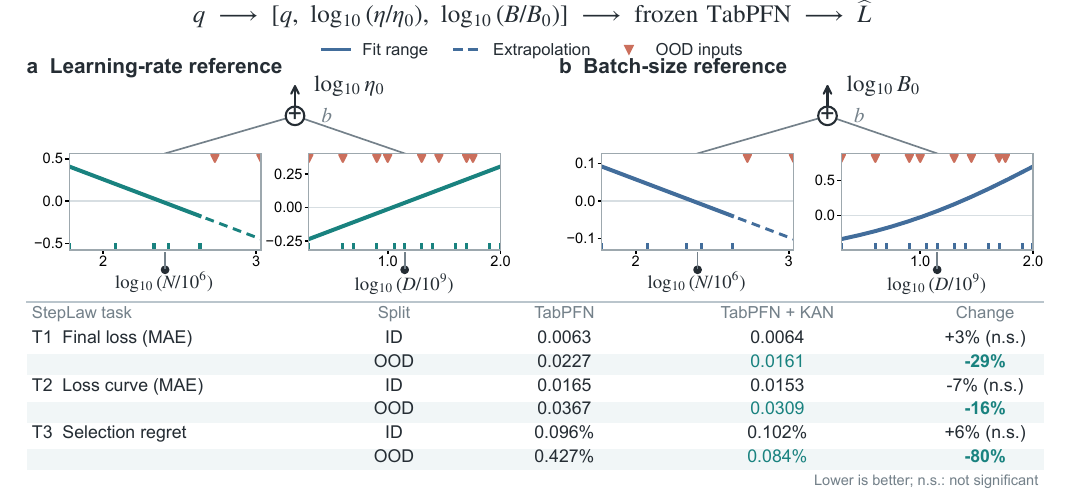}
\caption{\textbf{Learned coordinates improve prediction and selection at larger scales.} (a,b) Archived seed-0 KAN edge functions, centered for display; each pair sums with an intercept to a reference value. Bottom: StepLaw results against the default, untuned TabPFN route, which is Table~\ref{tab:sec2_summary}'s TabPFN predictor. MAE is in nats, regret in percent. TabPFN pools corpora; the KAN route uses StepLaw context. Matched-context controls and uncertainty estimates are in Appendix~\ref{app:kan_coordinates}.}
\label{fig:kan}
\end{figure}

\textbf{Learning a representation of training choices.}
\label{sec:better_correction}
Two small KANs learn reference values from the best-performing runs within each StepLaw training group, as functions of model size and token budget (Figure~\ref{fig:kan}). We append log-ratios to these references to the original configuration fields. TabPFN's backbone stays frozen, and no queried-run outcomes enter the KAN fit.

\textbf{Improving predictions and training decisions.}
Against the default TabPFN route, larger-model final-loss and curve errors decrease by $29\%$ and $16\%$. Selecting the configuration with lowest predicted loss reduces OOD regret by $80\%$: references learned from final-loss sweeps also help forecast trajectories and make decisions. ID changes are inconclusive. OOD gains persist against a same-context baseline (Appendix~\ref{app:kan_coordinates}).

\textbf{Inspecting what the KAN learns.}
The learning-rate reference decreases with model size and increases with token budget; batch size varies nonlinearly with token budget. These inspectable relationships are fitted, not universal training laws.

We next test the backbone against prospectively arriving outcomes.

\section{A prospective forecast of the Marin Hero Run}
\label{sec:marin_hero}

Can the same meta-model incorporate observations from an ongoing run and forecast what follows? We test this on the Marin Hero Run, a public MoE model with approximately 535B total and 23B active parameters \citep{hall2026hero}. On 25 August 2026, we froze a completion forecast (V0) using four smaller ladder runs and five hero observations through $3.84\%$ of training, its \emph{prefix}. Compute, progress, and active parameter count specify each future Paloma macro-loss query. Unlike Section~\ref{sec:sec2_dynamics}, target-run observations enter the context; backbone weights remain unchanged.

\begin{figure}[!htbp]
\centering
\includegraphics[width=\linewidth]{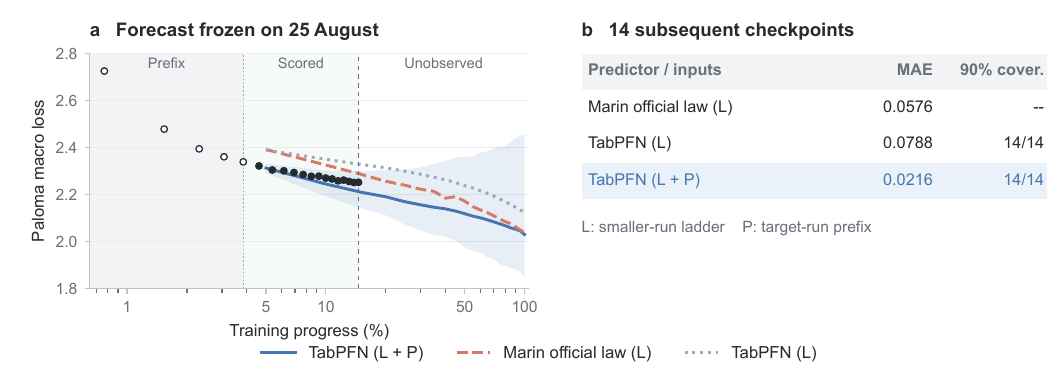}
\caption{\textbf{Forecast first, score as observations arrive.} (a) The V0 forecast frozen on 25 August: open/filled points are supplied/subsequent observations; the blue band is TabPFN's nominal $90\%$ interval. Scoring ends at $14.61\%$. (b) MAE and nominal $90\%$ coverage on the 14 subsequent checkpoints. L denotes the smaller-run ladder, P the target-run prefix; the official law provides no interval. Target: dropless Paloma macro loss. Full chronology and scoring: Appendix~\ref{app:hero_protocol}.}
\label{fig:marin}
\end{figure}

Our reference is Marin's official scaling law, which extrapolates the same four ladder runs and was archived alongside V0. On all 14 subsequent checkpoints available in the 24 September public-log check, V0's MAE is $0.0216$, versus $0.0576$ for the official law (Figure~\ref{fig:marin}), and its nominal $90\%$ interval covers all 14 observations. The gain comes from conditioning on the ongoing run: given only the ladder records, TabPFN's MAE is $0.0788$.

The forecast was preserved and never overwritten by later predictions. V0 preceded subsequent evaluations, but scoring rules were finalized after the first two arrived: predictions were frozen prospectively, not the entire evaluation protocol. The two predictors also use different information, since the official law does not see the prefix. This case extends a frozen backbone from offline queries to forecasts conditioned on an ongoing run; it does not isolate a benefit from ladder records beyond the prefix alone.


\section{Discussion and limitations}
\label{sec:discussion}

\textbf{Training questions can share a predictive backbone.} Configuration fields, prediction targets, and contextual observations can change without new TabPFN weights. This extends meta-modeling beyond a fixed configuration-to-loss mapping, while retaining task-specific representations and target processing.

\textbf{From reuse to reliable decisions.} Three challenges remain: representing genuinely new regimes, such as unseen data recipes; recognizing unreliable extrapolation; and demonstrating better experimental decisions. Our offline selection results motivate prospective, equal-budget comparisons counting pilot runs, search, and adaptation. Lower prediction error alone does not establish compute savings.

\section{Related work}
\label{sec:related_work}

\textbf{Scaling laws and training dynamics.}
Scaling laws extend beyond model and data size to mixtures \citep{ye2025mixinglaws}, sparsity \citep{abnar2025sparsity}, and learning-rate schedules \citep{tissue2024annealing,luo2025multipower}. Hyperparameter transfer \citep{yang2021tensor,wang2024adamw} and scale-dependent tuning \citep{porian2024discrepancies} further qualify these predictions. Our numerical response functions complement, rather than replace, such structured laws.

\textbf{Learned predictors of training outcomes.}
OmniPred trains language models as general regressors \citep{song2024omnipred}; NCPL specializes an LLM to training outcomes \citep{ncpl2026}. Our question is whether a frozen numerical backbone can serve changing configuration fields, targets, and contexts, not whether text-based regression is inherently incapable of doing so.

\textbf{Learning curves and numerical in-context learning.}
Bayesian neural surrogates \citep{klein2017learning} and Deep Power Laws \citep{kadra2023dpl} connect curve prediction to optimization. PFNs amortize inference through pretraining \citep{muller2022transformers}; LC-PFN, FT-PFN, and Adam-PFN specialize it to curves and hyperparameter search \citep{adriaensen2023lcpfn,rakotoarison2024ifbo,athanasiadis2025adampfn}. We instead reuse TabPFN's general tabular prior \citep{hollmann2023tabpfn,hollmann2025tabpfn} across training questions, without pretraining a dedicated training-curve model.

\textbf{Downstream prediction.}
Model ladders predict through intermediate losses \citep{bhagia2025ladders}; observational laws use a low-dimensional capability space \citep{ruan2024observational}, and NeuNeu learns from accuracy trajectories and token-level losses \citep{hu2026neuneu}. Predictability remains setting-dependent \citep{lourie2025unreliable}.

\textbf{Inspectable adaptation.}
KANs learn univariate edge functions \citep{liu2024kan}. We use them to learn inspectable configuration coordinates for a frozen predictor, rather than replace its backbone or establish causal training laws.

\section*{Reproducibility statement}
Experiments use public run corpora and released checkpoints of \citet{ncpl2026}. Splits, information sets, selection rules, and leakage checks for the reported results are documented in \S\ref{sec:shootout}--\S\ref{sec:marin_hero} and Appendices~\ref{app:experimental_setup}--\ref{app:further_analyses}. Code, cached predictions, and the timestamped forecast protocol will be released upon acceptance.

\section*{AI use statement}
The authors used generative AI assistants (OpenAI ChatGPT, Codex-based coding agents, and Anthropic Claude via Claude Code) to help refine research hypotheses and experimental designs; implement and refactor code; run and monitor experiments; build audit procedures; analyze results; design figures; search the literature; and draft, translate, and edit the manuscript. The research questions were posed by the authors, and all final methodological choices and interpretations of results were made by the authors; every experiment was approved by the authors and executed under their supervision. The authors inspected raw outputs and reviewed the code, citations, figures, and claims in the submission. All reported measurements come from the deterministic evaluators and intervention procedures described in the paper, not from language-model judgments. The authors reviewed all AI-assisted content and take full responsibility for the final paper.

\bibliography{iclr2027_conference}
\bibliographystyle{plainnat}

\clearpage
\appendix
\noindent
Appendix~\ref{app:experimental_setup} gives experimental details; Appendices~\ref{app:additional_experiments}--\ref{app:forecast_results} supplement Sections~\ref{sec:shootout}--\ref{sec:marin_hero}; Appendix~\ref{app:further_analyses} covers transfer, field-order robustness, and training responses.

\section{Experimental details}
\label{app:experimental_setup}
\subsection{Datasets, prediction tasks, and splits}
\label{app:sec2_inventory}
Table~\ref{tab:dataset-structure} lists the corpora. T1 predicts final loss from configuration; T2 adds progress to predict checkpoint loss without queried-run observations; T3 selects configurations by predicted loss. Marin and StepLaw keep the released train/ID/OOD splits of \citet{ncpl2026}; ID is used for selection. Targets keep a fitted corpus-specific Chinchilla offset. Main-table selection is described in Appendix~\ref{app:t1notes}. Released NCPL has no expert or recipe fields; its one-seed MoE fine-tune keeps the original schema.

\begin{table}[H]
  \caption{\textbf{Corpora represented in the common schema.}
  Counts are frozen source records before task-specific eligibility; the evaluated
  subset is stated where it differs. Records pair configuration metadata with
  scalar outcomes or checkpoint trajectories. Task-specific inputs differ;
  the proposed typed schema is documented separately in Appendix~\ref{app:training_record_schema}.}
  \label{tab:dataset-structure}
  \centering
  \small
  \setlength{\tabcolsep}{4pt}
  \renewcommand{\arraystretch}{1.08}
  \begin{tabularx}{\linewidth}{@{}lr>{\raggedright\arraybackslash}p{1.22in}>{\raggedright\arraybackslash}X@{}}
    \toprule
    Corpus & Runs & Role in this paper & Principal variation and data style \\
    \midrule
    Marin & 2,549 & benchmark
      & Dense-LM $N,D$, optimizer and decay sweeps; exponent-coded $\epsilon$ and step warmup. \\
    StepLaw & 2,581 & benchmark
      & Learning-rate$\times$batch sweeps across $N,D$; non-embedding parameter count and smoothed train loss. \\
    Marin-MoE & 1,036 & benchmark
      & Expert count and top-$k$; active and total parameters; fractional warmup. \\
    DataDecide & 1,100 & benchmark (950 evaluated)
      & 25 data recipes crossed with model size; categorical recipe identity and checkpoint losses. \\
    Delphi & 88 & scaling extrapolation
      & Iso-FLOP runs and a three-rung held-out ladder; Paloma macro-loss trajectories. \\
    Delphi-HP & 3,737 & prediction / selection
      & Seven optimizer hyperparameters in 24 width$\times$batch$\times$token cells. \\
    \midrule
    \textbf{Total} & \textbf{11,091} & \multicolumn{2}{l@{}}{\textbf{6 corpora; heterogeneous fields, clocks, units and readouts.}} \\
    \bottomrule
  \end{tabularx}
\end{table}

\paragraph{DataDecide.}
DataDecide \citep{magnusson2025datadecide} crosses 25 recipes with 14 OLMo \citep{groeneveld2024olmo} sizes (4M--1B) at $D=100N$. After removing early-stopped auxiliary seeds, 950 runs and 19,124 checkpoints remain, scored as log C4 perplexity. The 750M and 1B runs are OOD; the rest is split 8:2 by recipe--size group.

\subsection{Baseline fitting and statistical evaluation}
\label{app:evaluation}
Parametric baselines fit $L=E+A/N^\alpha+B/D^\beta$ \citep{hoffmann2022chinchilla} on training records, optionally with a $\log\eta$ term; XGBoost \citep{chen2016xgboost} is tuned on ID. The MoE sparsity-aware law \citep{abnar2025sparsity} uses $(N_{\rm tot},D,S)$.

\subsection{Evaluation and figure-construction protocols}
\label{app:sec2_capability_protocol}
Figure scripts verify source hashes, run identities, split disjointness, and target alignment. Illustrative trajectory examples use metadata rules (for example, the lower-median run of a fixed group).

\paragraph{Configuration prediction (Figure~\ref{fig:sec2_configurations}).}
Delphi-HP predicts every trial in each of 24 held-out cells (TabPFN: eight estimators, mean readout; baselines tuned by grouped five-fold CV on the remaining cells; losses capped at 6.0). Mean Spearman correlation is $0.967/0.875/0.820/0.735$ for TabPFN/XGBoost/quadratic ridge/kNN, and MAE $0.045/0.100/0.143/0.141$. Mixture prediction holds out one mixture at fixed family, size, and step, averaging seeds 0--2; the denominator is the lowest M1--M4 MAE in each cell (Appendix~\ref{app:mixlaws}).

\paragraph{Trajectories (Figure~\ref{fig:sec2_dynamics}).}
\label{app:sec22_contrasts}
Curves use joint-context, two-seed medians; displays omit progress $\le5\%$, but scores keep every checkpoint. For centered contrasts, StepLaw runs are grouped by all settings except learning rate (40 ID and 85 OOD groups, 4,219 pairs), and each pair's mean gap over shared times $T_{ij}$ is removed:
\[
\widetilde{\Delta L}_{ij}(t)=L_i(t)-L_j(t)
-|T_{ij}|^{-1}\sum_{u\in T_{ij}}[L_i(u)-L_j(u)].
\]
OOD contrast MAE is $0.0572$ for any shared-curve predictor $g(t)+b_i$, $0.0371$ for NCPL, and $0.0354$ for TabPFN. On Marin OOD (7,844 checkpoints, 415 runs), full/early ($\le5\%$)/later MAE is $0.0457/0.2419/0.0264$ for TabPFN versus $0.0350/0.1423/0.0244$ for NCPL. MoE scores 341 checkpoints from 23 OOD runs: full/early/later MAE is $0.0326/0.0527/0.0318$ for TabPFN versus $0.1086/0.1671/0.1065$ for MoE-fine-tuned NCPL. The early and later subsets contain 12 and 329 checkpoints, respectively; the approximately $70\%$ reduction therefore persists beyond the early phase omitted from the example curves. Predictions are aligned by run identifier and recorded progress and checked against the source losses before computing these aggregates. The DataDecide recipe ablation gives OOD curve MAE $0.0133$ with the recipe field, $0.1027$ without, and $0.0232$ for XGBoost.

\paragraph{Observables and downstream predictions (Figure~\ref{fig:sec2_downstream}).}
The observable panels use full-configuration gradient-norm predictions and numerical-geometry, scale-normalized parameter predictions (Appendix~\ref{app:obs}). The two runs are the lower-median lexicographic names within 130M MARS ID and 520M Mini OOD. Both panels show all checkpoints without smoothing, and dashed curves average two seed-specific median predictions. The portable plot data record exact run IDs, raw-prediction hashes, and the selection rule. DataDecide uses the historical-checkpoint protocol in Appendix~\ref{app:ddecision} and displays all 50 OOD size--recipe groups, averaging run seeds and TabPFN seeds for each point. Its reported MAE instead averages absolute errors per prediction seed and then by size--recipe group; it is not the error of the plotted average. Neither direct predictor receives target-run outcomes.

\paragraph{Selection (Figure~\ref{fig:sec2_selection}).}
Figure~\ref{fig:sec2_selection}a pairs history-only and history-plus-prefix MAE for all seven Delphi ladder runs, averaging errors over future checkpoints and three TabPFN seeds within each run; the diamonds then average runs equally (Appendix~\ref{app:delphi_prefix}). Green lines improve and coral lines worsen. Panel b shows the OOD sweep with 108 candidates at $N=537$M, $D=50$B. It displays seed 0. Its contours interpolate held-out losses for display only and do not enter selection. Panel c uses original, unclipped Paloma loss to score the selected candidate in each of 23 eligible Delphi cells; diamonds are means, boxes span the 25th--75th percentiles, and internal lines mark medians. Panel d averages the 20 initializations within each condition, then weights twelve OOD conditions equally. Shading is a pointwise percentile 95\% interval from 20,000 condition-level bootstrap draws, not an interval over seeds (Appendix~\ref{app:bo_feedback}). Symmetric-log axes have a linear region below $0.01$ nats in (c) and $0.01\%$ regret in (d); zero remains visible. Portable inputs contain source hashes and the frozen matched-feedback protocol.

\subsection{Adaptation and inference cost}
\label{app:cost}
\begin{table}[!htbp]
\centering
\small
\caption{Adaptation and prediction cost. NCPL training entries are estimates; timings are medians of three warm runs and exclude foundation-model pretraining. They concern the joint-context benchmark, not the main-table final-loss routes.}
\label{tab:cost}
\begin{tabular}{lrr}
\toprule
Operation & Final loss & Curves \\
\midrule
NCPL fine-tuning, FLOP estimate (GPU-h) & 7.0 & 44.3 \\
NCPL fine-tuning, throughput estimate (GPU-h) & 24.4 & 145.8 \\
NCPL inference, one H200 fp32 (s) & 15.10 & 274 \\
NCPL inference, TF32 allowed (s) & 3.44 & 61.14 \\
TabPFN context processing + prediction, H200 (s) & 1.79 & 12.64 \\
TabPFN context processing + prediction, CPU (s) & 19.46 & 1,296 \\
XGBoost selected fit + prediction, CPU (s) & 0.19 & 0.86 \\
XGBoost 12-configuration search, CPU (s) & 1.95 & 5.45 \\
\bottomrule
\end{tabular}
\end{table}

Final prediction uses 3,225 context rows and 1,905 queries; curves use 46,595 context points and 28,743 queries (TabPFN: eight estimators). NCPL training cost is extrapolated either from measured step time on H200s (throughput) or from the NCPL fine-tuning recipe \citep{ncpl2026} for Qwen3-1.7B \citep{yang2025qwen3} at 40\% utilization (FLOPs). The 679-configuration TabPFN tuning search costs 9.9 GPU-hours (Appendix~\ref{app:tune}).

\FloatBarrier

\section{Additional results for TabPFN as a meta-model}
\label{app:additional_experiments}
\subsection{Delphi-HP: prediction and selection on held-out conditions}
\label{app:delphi_hp}
The 3,737-trial sweep \citep{held2026delphi} has 24 width--batch--token cells and seven log-transformed optimizer hyperparameters (two learning rates, epsilon, clipping, z-loss, and two betas). One cell is withheld at a time. Ranking uses loss clipped at 6.0; regret uses raw final Paloma loss on the 23 cells with at least 150 trials. TabPFN regret is $0.0038$ $[0.0024,0.0052]$ versus XGBoost's $0.0254$. An equal-candidate-count check draws 61 trials per cell, using exactly the same candidates for every method and repeating the subsampling 100 times. Averaging repeats within each of all 24 cells, then cells equally, gives $0.00383$ versus $0.01699$. No trial overlaps with Marin.

\subsection{Mixture-law fitting and sensitivity}
\label{app:mixlaws}
We fit M1--M4 of \citet{ye2025mixinglaws} and query TabPFN on identical leave-one-mixture-out folds (268 Pythia \citep{biderman2023pythia} runs in four families), at each cell's largest step shared by all mixtures. Errors average seeds 0--2; paired mixture bootstraps ($B=2000$) compare with M4. On 5Mix, TabPFN gives $0.0084$--$0.0113$ nats versus M1's $0.0133$--$0.0147$ and M4's $0.0271$--$0.0286$; every ratio interval against M4 excludes one. 3Mix is close ($0.0253$ versus M1's $0.0266$). M4 is $2.2$--$2.8\times$ better on 2Mix, and M2 leads at three of four RedPajama sizes, where some M1/M4 folds diverge.

\subsubsection{Main-table selection and prediction settings}
\label{app:t1notes}
Table~\ref{tab:sec2_summary} selects the lowest ID-MAE row for each method and corpus, separately for final loss and curves, and retains that row's other metrics; neither OOD scores nor individual metric optima determine selection. TabPFN candidates include only the archived zero-shot mean/median and joint/per-corpus variants, excluding the doc-guided tuning ladder. Their unrounded, prediction-cache-audited ID MAEs determine selection; the two parametric laws are grouped together, while the sparsity-aware law remains separate. Cross-lab TabPFN is excluded because it lacks in-corpus context. NCPL paper and reproduction rows remain separate. For TabPFN final loss, the selected dense-model rows are per-subset median on Marin and joint mean on StepLaw; curves use joint median on both. DataDecide uses mean for final loss and median for curves, and MoE uses median for both. In StepLaw, the joint-mean, joint-median, and per-subset-median final-loss ID MAEs all round to $0.0063$, but their unrounded values are $0.00627644$, $0.00632527$, and $0.00634659$, respectively; joint mean is therefore selected. Zero-shot here denotes no task-specific fitting of TabPFN weights, not absence of labeled context or ID-based inference-variant selection.

\FloatBarrier
\subsection{DataDecide: recipe information and context controls}
\label{app:datadecide}
\begin{figure}[htbp]
\centering
\includegraphics[width=0.82\linewidth]{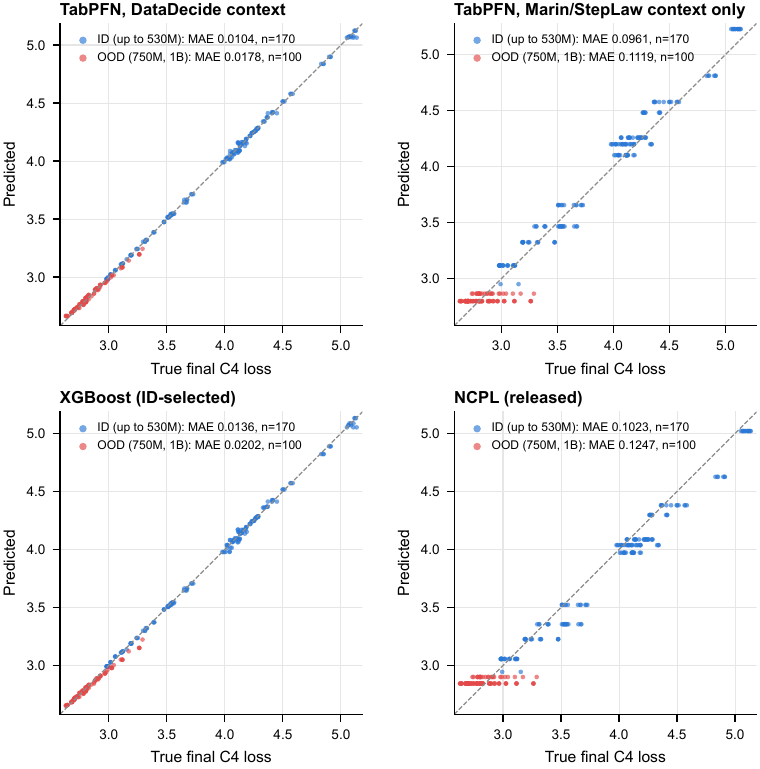}
\caption{\textbf{DataDecide final-loss prediction}, predicted versus true C4 loss (blue: ID sizes up to 530M; red: held-out 750M/1B). Top left: TabPFN with DataDecide training context (OOD MAE $0.0178$). Top right: TabPFN with only Marin/StepLaw context, where recipe levels have no supporting examples ($0.1119$). Bottom: ID-selected XGBoost ($0.0202$) and released NCPL, whose interface has no recipe field ($0.1247$). TabPFN uses seed 0 and the mean readout.}
\label{fig:datadecide}
\end{figure}

Without recipe identity, OOD final-loss MAE rises from $0.0178$ to $0.1117$. Figure~\ref{fig:datadecide} shows why: with DataDecide records as context, TabPFN tracks recipe-dependent losses at every size, whereas context without DataDecide recipes, or NCPL's recipe-free interface, collapses predictions into size-quantized bands.

 Released NCPL inputs are identical across recipes at a given size, forcing identical predictions. Size explains $97.6\%$ of total variance, but recipe explains $99.2\%$ of within-size variance, so recipe context is necessary for choosing data at fixed scale.

\subsection{Predicting observables and their trajectories}
\label{app:obs}
\label{app:sec2_observables}
Public Marin logs supply internal observables; each target has its own eligible context and preprocessing. The backbone remains frozen, with median readouts. We report the norm evaluation used in the main text.

\paragraph{Norm targets and protocol.}
Both norm targets are sums of the L2 norms of 291 disjoint tensors, not the global L2 norm of a concatenated vector. We use this tensor-sum definition consistently across all runs. The fixed split contains 281 training, 43 ID, and 233 OOD runs. Curve contexts contain 64 checkpoints per training run (17,984 rows); evaluation retains all 761,981 held-out timestamps, including initialization. Queries supply configuration and progress, never the target run's weights, initial measured norm, or observed prefix. We average errors within each run, then across runs and two prediction seeds (eight estimators per seed).

Parameter-norm prediction preserves numerical types for model-geometry fields and predicts $\log_{10}S(q,t)-\log_{10}\widehat S_0(q)$, where $\widehat S_0$ is fitted from training-run widths and initial norms. Test queries use only configuration to estimate this scale; the backbone is not fine-tuned. Gradient-norm predictions use the full-configuration model without this additional normalization.

\begin{table}[htbp]
\centering
\caption{\textbf{Observable predictions.} Run-macro $\log_{10}$ MAE, averaged over two seeds. Final-value predictors are fitted separately, not read from the curve endpoint. All variants use the same tensor-sum targets. Parameter variants compare default feature typing, explicit numerical geometry, and additional scale normalization.}
\label{tab:corrected_observables}
\small
\begin{tabular}{lrrrr}
\toprule
 & \multicolumn{2}{c}{Full curve} & \multicolumn{2}{c}{Final value} \\
 & ID & OOD & ID & OOD \\
\midrule
Parameter: default feature typing & 0.0363 & 0.1615 & 0.0290 & 0.1973 \\
Parameter: numerical geometry & 0.0368 & 0.1149 & 0.0288 & 0.1228 \\
Parameter: + scale normalization & 0.0341 & 0.0726 & 0.0275 & 0.0868 \\
Gradient: full configuration & 0.0463 & 0.0994 & 0.0240 & 0.1026 \\
\bottomrule
\end{tabular}
\end{table}

\begin{figure}[htbp]
\centering
\includegraphics[width=\linewidth]{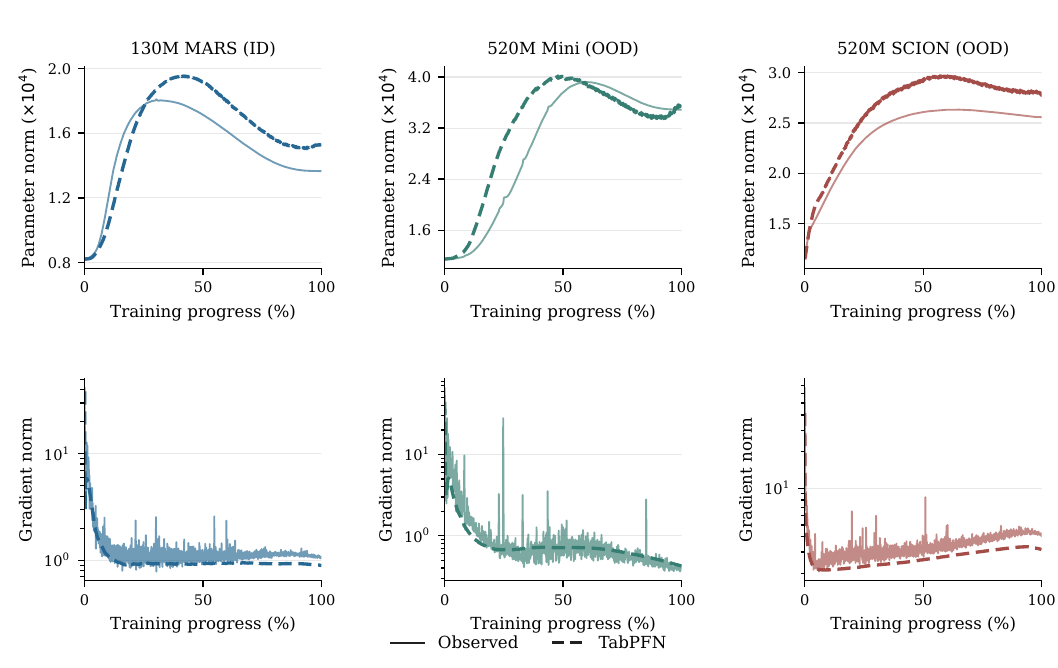}
\caption{\textbf{Internal-observable trajectories.} The first two columns expand the runs in Figure~\ref{fig:sec2_downstream}; the third adds a 520M SCION run. Each is the lower-median lexicographic run within its size--optimizer--split group. Every recorded checkpoint is shown, with a log scale for gradients. Solid curves are observations and dashed curves average two TabPFN median predictions. These illustrative runs are not selected by error or claimed to be typical; Table~\ref{tab:corrected_observables} scores all held-out runs.}
\label{fig:sec2_trajectories}
\end{figure}

The normalized parameter model improves curve error on 212 of 233 OOD runs relative to default feature typing. We use one fixed variant for all displayed parameter predictions, not a test-selected variant per optimizer. Figure~\ref{fig:sec2_trajectories} includes full trajectories and an additional optimizer; aggregate metrics retain spikes and all initial checkpoints.

\subsection{DataDecide: downstream prediction from training histories}
\label{app:ddecision}
The DataDecide evaluation release \citep{magnusson2025datadecide} ends earlier than its perplexity archive (89\% of planned training at 530M, 41\% at 750M), so the target is the last available evaluation, not necessarily completed training. The fixed split has 680 training, 170 ID, and 100 OOD (750M/1B) runs. Targets are ten OLMES \citep{gu2024olmes} accuracies and their macro average. We score each size--recipe group equally, first averaging absolute errors over run seeds and then over two TabPFN prediction seeds. The 1B evaluation has three run seeds per recipe; 750M has one.

\paragraph{Historical-checkpoint protocol used in the main text.}
The earlier endpoint-only context covered training progress of approximately $89$--$100\%$, despite the 750M queries occurring at $41\%$. We now retain multiple positive-step checkpoint evaluations from the same training runs, selected at evenly spaced time indices without consulting target values. To choose among caps of 8, 32, and all available checkpoints, we use only original-training runs: successively train on smaller sizes and query 150M, 300M, and 530M runs near $40\%$ progress and at their last evaluation. Macro-accuracy MAE, equally weighted across folds and recipe--progress groups, is $1.451$, $1.676$, and $1.669$ percentage points, respectively. We lock the eight-checkpoint cap before final evaluation, yielding 4,975 context rows from 680 runs. No observations from an ID or OOD query run enter this context.

All configuration and progress fields retain numerical types; recipe is categorical. Each target uses the same selected rows and a frozen TabPFN v3 backbone (eight estimators; seeds 0 and 1; median readout), with no target-specific neural fine-tuning. XGBoost receives the same rows and features. Its per-target grid comprises depths $\{3,6\}$, learning rates $\{0.05,0.1,0.3\}$, and tree counts $\{300,800\}$, selected by grouped ID MAE. Its ID scores are therefore tuning performance.

\begin{table}[htbp]
\centering
\caption{\textbf{Historical observations improve downstream prediction.} Macro-accuracy MAE in percentage points. The new methods use identical historical-checkpoint contexts. Targets and held-out runs are unchanged from the endpoint protocol. XGBoost ID error is its grid-selection score.}
\label{tab:downstream_history}
\small
\begin{tabular}{lrrrr}
\toprule
Predictor / context & ID & 750M & 1B & OOD average \\
\midrule
TabPFN / endpoints (earlier) & 0.66 & 1.52 & 3.49 & 2.51 \\
TabPFN / historical checkpoints & 0.64 & 0.78 & 2.46 & 1.62 \\
XGBoost / historical checkpoints & 0.70 & 1.26 & 5.53 & 3.40 \\
\bottomrule
\end{tabular}
\end{table}

\begin{figure}[htbp]
\centering
\includegraphics[width=\linewidth]{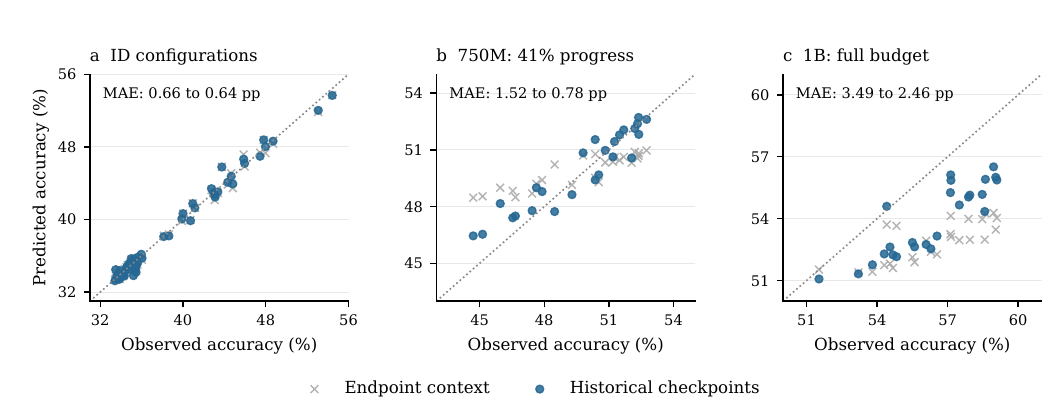}
\caption{\textbf{Downstream prediction by evaluation regime.} Earlier endpoint-context predictions (gray) and revised historical-checkpoint predictions (blue), with unchanged query outcomes. Points average run and prediction seeds within size--recipe groups; MAE is computed before averaging predictions. ID fit is already close, while additional historical observations improve both OOD regimes without fully removing the 1B bias.}
\label{fig:downstream_history}
\end{figure}

\paragraph{Improvement and remaining extrapolation error.}
The revised protocol lowers OOD macro error by $35\%$ relative to the earlier endpoint protocol and lowers error on all ten constituent targets (Table~\ref{tab:downstream_tasks}). It does not eliminate scale-dependent bias: 1B predictions remain low by $2.44$ percentage points on average (Figure~\ref{fig:downstream_history}). In the original context, $84\%$ of 1B outcomes exceeded the largest training accuracy ($54.46\%$). The new experiment adds temporal evidence, not larger training models. These results support better use of existing records, not universally calibrated scale extrapolation.

Layer and head counts are treated as numerical features rather than inferred categories.

The constituent benchmarks are ARC (Challenge and Easy; \citealp{clark2018arc}), BoolQ \citep{clark2019boolq}, CommonsenseQA \citep{talmor2018commonsenseqa}, HellaSwag \citep{zellers2019hellaswag}, MMLU \citep{hendrycks2020mmlu}, OpenBookQA \citep{mihaylov2018openbookqa}, PIQA \citep{bisk2019piqa}, SocialIQA \citep{sap2019socialiqa}, and WinoGrande \citep{sakaguchi2019winogrande}.

\begin{table}[htbp]
\centering
\caption{\textbf{Constituent downstream targets.} OOD MAE in percentage points, equally weighted across size--recipe groups. The two history methods use identical contexts; the endpoint column records the earlier TabPFN protocol. These are descriptive comparisons across the same fixed test split.}
\label{tab:downstream_tasks}
\small
\begin{tabular}{lrrr}
\toprule
Target & TabPFN: endpoints & TabPFN: history & XGBoost: history \\
\midrule
ARC-Challenge & 2.12 & 1.43 & 3.40 \\
ARC-Easy & 1.87 & 1.73 & 4.11 \\
BoolQ & 6.85 & 5.52 & 6.53 \\
CommonsenseQA & 2.82 & 2.46 & 4.78 \\
HellaSwag & 2.82 & 1.69 & 6.11 \\
MMLU & 1.16 & 0.54 & 1.71 \\
OpenBookQA & 2.62 & 1.92 & 3.54 \\
PIQA & 1.58 & 1.10 & 2.21 \\
SocialIQA & 1.80 & 1.25 & 2.26 \\
WinoGrande & 2.92 & 1.33 & 2.78 \\
Macro accuracy & 2.51 & 1.62 & 3.40 \\
\bottomrule
\end{tabular}
\end{table}

\FloatBarrier
\subsection{Loss-curve examples on Marin}
\label{app:marin_curves}
\begin{figure}[htbp]
\centering
\includegraphics[width=0.92\linewidth]{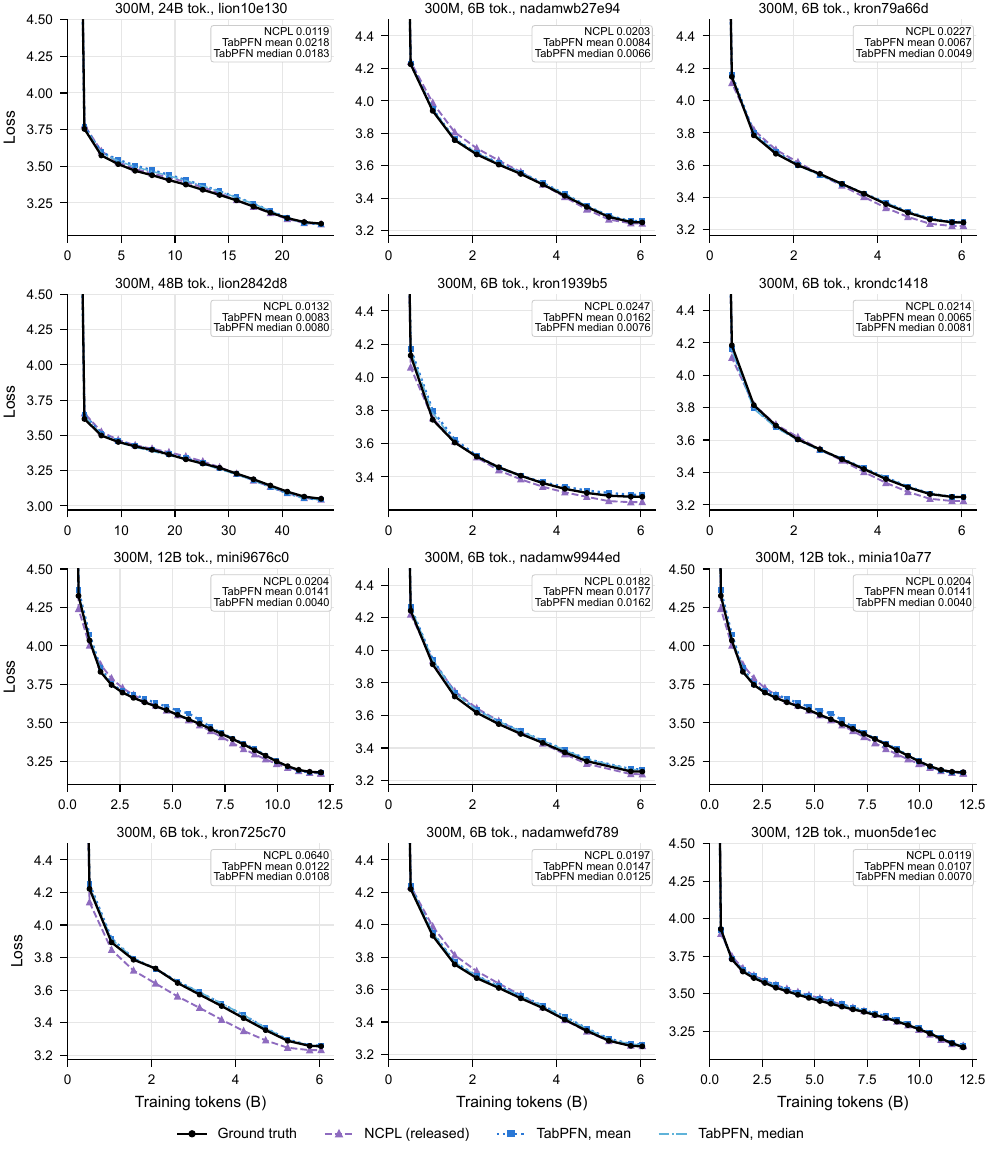}
\caption{\textbf{Loss-curve predictions for the 12 Marin ID runs (300M) sampled in Figure~13 of \citet{ncpl2026}.} Black: ground truth; purple dashed: released NCPL; blue dotted and dash-dotted: TabPFN mean and median. Boxes give per-run MAE.}
\label{fig:curves_id12}
\end{figure}

\begin{figure}[htbp]
\centering
\includegraphics[width=0.92\linewidth]{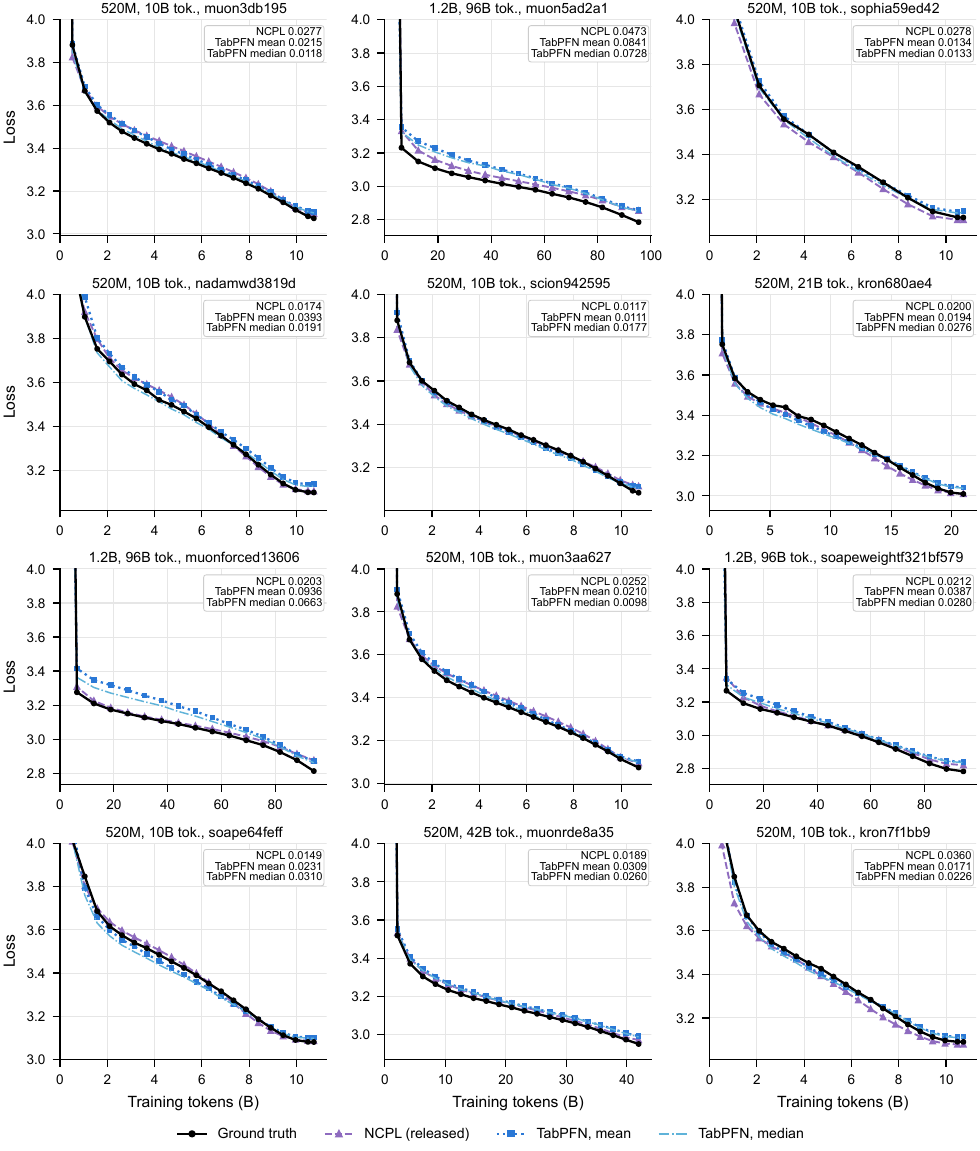}
\caption{\textbf{Loss-curve predictions for the 12 Marin OOD runs of Figure~14 of \citet{ncpl2026}.} Lines and boxes as in Figure~\ref{fig:curves_id12}.}
\label{fig:curves_ood12}
\end{figure}

\begin{figure}[htbp]
\centering
\includegraphics[width=\linewidth]{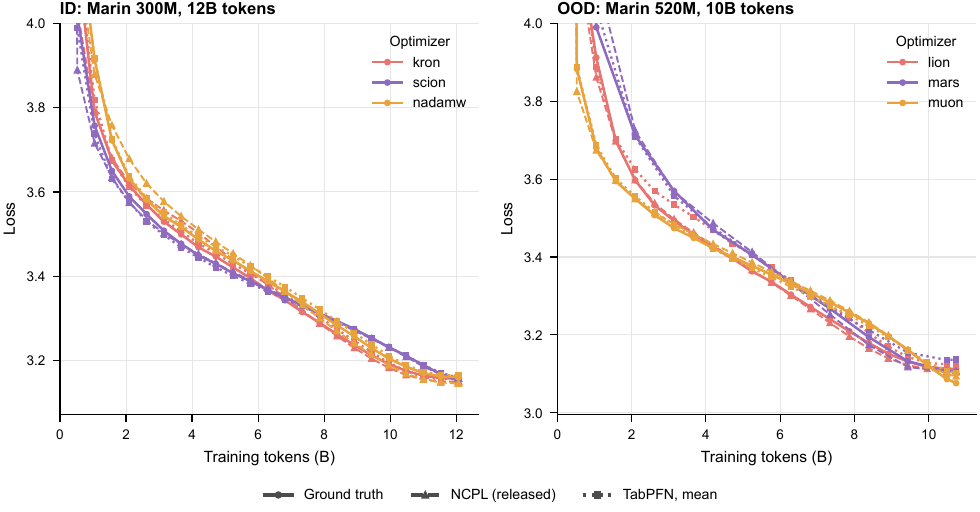}
\caption{\textbf{Optimizer-specific loss curves in the setting of Figure~15 of \citet{ncpl2026}}: the median-final-loss run of each $(N,D,\text{optimizer})$ group. Solid: ground truth; dashed: NCPL; dotted: TabPFN mean. Left: ID; right: OOD.}
\label{fig:curves_optim}
\end{figure}

Figures~\ref{fig:curves_id12}--\ref{fig:curves_optim} repeat the three curve displays of \citet{ncpl2026} (their Figures 13--15) on the same run names, adding TabPFN. These use the default TabPFN v3 checkpoint with two estimators and seed 0 (mean and median readouts) rather than the joint-median route of Table~\ref{tab:sec2_summary}; NCPL is the released checkpoint, and no target-run loss is supplied to either model. Boxes in the first two figures give each run's MAE.

\FloatBarrier
\subsection{Updating forecasts with observations from the target run}
\label{app:delphi_prefix}
This uses Delphi's 88-run scaling suite (Appendix~\ref{app:delphi_meta}): 81 historical runs and seven held-out ladder runs at $10^{21}$--$10^{23}$ FLOPs. At cutoff $p$, the target run's observations with progress $f\le p$ are appended to the historical context, and all later checkpoints are scored. Both TabPFN arms use eight estimators, seeds 0--2, mean readout, and a fixed Chinchilla offset fitted on historical runs; only the context changes. Errors are averaged within runs, then across runs.

\begin{table}[htbp]
\centering
\small
\caption{\textbf{New target-run observations can improve future-loss predictions.} Run-averaged remaining-curve MAE (nats) on the seven Delphi ladder runs. Columns score different future windows; the median row is not a matched control.}
\label{tab:delphi_prefix}
\begin{tabular}{lrrr}
\toprule
Observed target prefix & $20\%$ & $50\%$ & $80\%$ \\
\midrule
TabPFN mean, history only & .05787 & .05425 & .06344 \\
TabPFN mean, history + prefix & .02537 & .03078 & .03922 \\
Relative MAE reduction & $56.2\%$ & $43.3\%$ & $38.2\%$ \\
Runs improved & 4/7 & 3/7 & 4/7 \\
\midrule
TabPFN median, history + prefix & .02378 & .02377 & .03949 \\
XGBoost, history + prefix & .02367 & .01951 & .01751 \\
Power law, prefix with fixed $E$ & .08158 & .07808 & .11555 \\
Last observed value & .17290 & .11438 & .16293 \\
\bottomrule
\end{tabular}
\end{table}

\subsection{Configuration selection and context updates}
\label{app:bo}
\begin{figure}[htbp]
\centering
\includegraphics[width=\linewidth]{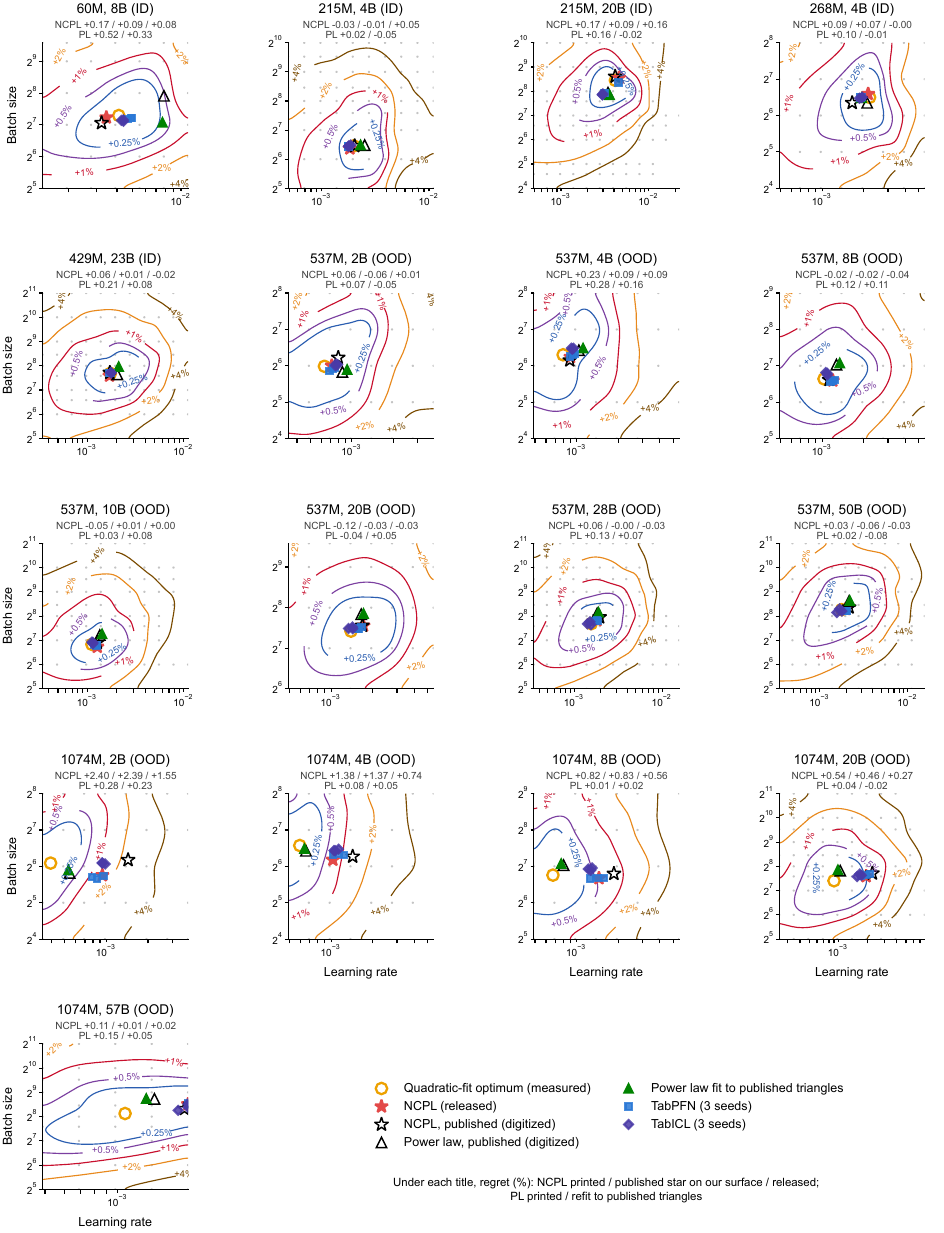}
\caption{\textbf{All 17 held-out StepLaw sweeps.} Contours: smoothed measured loss relative to the sweep optimum; gray dots: candidates. Filled markers: optimum of a quadratic fit to the measured losses (orange circle), released NCPL (red star), power law fit to the published selections (green triangle), TabPFN (blue squares, three seeds), and TabICL \citep{qu2026tabiclv2} (purple diamonds). Hollow black markers: NCPL (star) and power-law (triangle) selections digitized from \citet{ncpl2026}. The two lines under each title give relative regret (\%): NCPL printed / published star evaluated on our surface / released checkpoint, and power law (PL) printed / refit.}
\label{fig:hpo_overlay}
\end{figure}

Figure~\ref{fig:hpo_overlay} shows all 17 held-out StepLaw sweeps used for one-shot selection (5 ID, 12 OOD) with each method's chosen candidate, together with NCPL's and the power law's published selections digitized from \citet{ncpl2026}. Our optimum coincides with the published one in every panel. At $N=1074$M and small $D$, the published NCPL choice lies at a larger learning rate than the released checkpoint's, which explains the difference between printed and reproduced NCPL regret.

\paragraph{Matched context-update controls.}
\label{app:bo_feedback}
Figure~\ref{fig:sec2_selection}d isolates the use of target-condition observations with three policies sharing a frozen TabPFN v3 backbone (eight estimators, mean readout). Historical context contains 3,225 Marin and StepLaw training rows. All policies use the same configuration features, fixed corpus-specific Chinchilla residual target, candidate order, and two random initial observations for each of 20 seeds and 17 sweeps. The seed controls both initialization and TabPFN preprocessing. A history-only policy never receives target outcomes; an initial-frozen policy receives the two initial outcomes but no later ones; an updating policy appends each observed outcome before making the next choice. All recommend the untested candidate with lowest predicted loss, breaking ties by original candidate order. Each prediction call queries the same complete candidate feature matrix; already evaluated candidates are masked only at selection. No unseen target loss enters preprocessing or the predictor. Initial observations count toward the total budget of twelve.

All three arms are rerun under this common protocol, producing 340 matched cases and 1,020 policy trajectories. Per-step prediction vectors and context indices verify each selection; regrets are recomputed from the original measured losses. Seeds are averaged within each condition; paired percentile bootstrap intervals resample the twelve conditions 20,000 times, not the 240 initializations as independent tasks.

\begin{table}[htbp]
\centering
\small
\caption{\textbf{Same-backbone feedback controls.} Mean best-found measured regret (\%) on twelve OOD conditions; budgets include two shared initial evaluations.}
\label{tab:bo_feedback}
\begin{tabular}{lrrrr}
\toprule
Context policy & $k=3$ & $k=5$ & $k=8$ & $k=12$ \\
\midrule
History only & .30200 & .13981 & .06210 & .01406 \\
History + initial observations, frozen & .16719 & .04733 & .00872 & .00220 \\
History + all revealed observations & .16719 & .03937 & .00160 & .00000 \\
\bottomrule
\end{tabular}
\end{table}

At budget five, conditioning on the initial outcomes reduces mean regret by $0.09249$ percentage points relative to history only (95\% interval $[0.01642,0.17977]$). Neither replay establishes prospective training-compute savings.

\FloatBarrier

\section{Learned representations and adaptation}
\label{app:adaptation_results}
\subsection{KAN-learned configuration coordinates}
\label{app:kan_coordinates}

\paragraph{Data and tasks.}
We use 1,517 StepLaw training runs to construct reference hyperparameters and prediction contexts. Final-loss prediction (T1) is evaluated on 370 ID and 694 larger-model OOD runs. Curve prediction (T2) uses their 5,177 and 10,745 recorded checkpoints, with 21,888 training checkpoints as context and no target-run prefix. All checkpoints of a run stay in one split. Configuration selection (T3) uses T1 predictions on five ID and twelve OOD model--data-size groups: we select the candidate with minimum predicted loss and report $100(L_{\mathrm{selected}}-L_{\min})/L_{\min}$, where $L_{\min}$ is the best measured candidate loss in that group.

\paragraph{Learning reference hyperparameters.}
For each of the 24 model--data-size groups in the training split, we average log learning rate and log batch size over its three lowest-loss runs. These summaries are the supervised targets of two separate pykan \citep{liu2024kan} $[2,1]$ networks, with standardized $(\log_{10}N,\log_{10}D)$ inputs and standardized targets. Each network has cubic splines with three grid intervals. We fit with 200 L-BFGS iterations, prune, refit for 100 iterations, simplify the edge functions symbolically, and refit for another 100 iterations without updating the grid.

The model-size edge is constrained to be linear in log space, an imposed power-law extrapolation prior. The data-size edge is chosen from $\{x,x^2,x^3,\tanh(x)\}$ by fit quality. A nonlinear candidate is retained only if leave-one-training-group-out RMSE is at least $5\%$ below the linear candidate's; this check uses only the 24 training groups and is applied once per target, before the prediction runs. It retains a linear edge for learning rate and a $\tanh$ edge for batch size. The KANs predict log reference values $s_\eta(N,D)$ and $s_B(N,D)$. The added configuration fields are
\begin{equation}
\Delta_\eta=\log_{10}\eta-s_\eta(N,D),\qquad
\Delta_B=\log_{10}B-s_B(N,D).
\label{eq:kan_coordinates}
\end{equation}
Original configuration fields, including learning rate and batch size, are retained. The same final-loss-derived references are used for curve prediction, whose queries also include training progress. No evaluation outcomes enter these fits or the contexts.

\paragraph{Default and matched-context comparisons.}
All predictors use the default TabPFN v3 checkpoint and eight estimators, with frozen backbone weights and the same fixed corpus-specific Chinchilla target offsets. The baseline is the default, untuned TabPFN route: joint Marin/StepLaw context, mean readout and seeds 0--2 for final loss, and median readout and seeds 0--1 for curves. Its final-loss and curve predictions are those of Table~\ref{tab:sec2_summary}. The KAN predictor is scored with these same task-specific readouts and seeds. T3 uses the corresponding T1 predictions without another readout selection. Figure~\ref{fig:kan} therefore reproduces Table~\ref{tab:sec2_summary}'s displayed StepLaw MAEs. The KAN route uses StepLaw-only context and added coordinates, so this is a comparison of two predictor routes, not an isolated coordinate ablation. Table~\ref{tab:kan_coordinates} also reports a StepLaw-only baseline without the coordinates, using the same readouts and seeds to isolate their effect. The KAN is fitted only to StepLaw; no Marin improvement is claimed.

\begin{table}[htbp]
\centering
\small
\caption{StepLaw results for the default TabPFN route and the KAN route: three seed-wise scores for T1/T3, two for T2, not an ensemble of predictions. T1/T2 report MAE in nats; T3 reports mean relative regret in percent.}
\label{tab:kan_coordinates}
\begin{tabular}{lrrrrrr}
\toprule
& \multicolumn{2}{c}{Final loss} & \multicolumn{2}{c}{Loss curve} & \multicolumn{2}{c}{Selection (\%)} \\
Predictor & ID & OOD & ID & OOD & ID & OOD \\
\midrule
TabPFN & 0.00628 & 0.02273 & 0.01653 & 0.03666 & 0.0960 & 0.4271 \\
TabPFN (StepLaw only) & 0.00640 & 0.02472 & 0.02671 & 0.04065 & 0.0694 & 0.3471 \\
TabPFN + KAN (StepLaw only) & 0.00644 & 0.01608 & 0.01531 & 0.03093 & 0.1018 & 0.0838 \\
\bottomrule
\end{tabular}
\end{table}

\paragraph{Uncertainty and scope.}
Paired bootstrap intervals use 10,000 resamples, with runs as units for T1/T2 (all checkpoints move together) and model--data-size groups for T3, after averaging seed-wise errors or regrets. Against the default TabPFN route, the OOD MAE differences are $-0.0066$ ($95\%$ interval $[-0.0076,-0.0057]$) for T1 and $-0.0057$ ($[-0.0072,-0.0044]$) for T2; OOD regret decreases by $0.343$ percentage points ($[-0.610,-0.117]$). ID intervals include zero for all three tasks: T1 $[-0.0004,0.0007]$, T2 $[-0.0026,0.0003]$, and T3 $[-0.080,0.097]$ percentage points. Against the matched StepLaw-context baseline, OOD mean errors and regret also improve, as shown in Table~\ref{tab:kan_coordinates}.

\paragraph{Figure provenance.}
Figure~\ref{fig:kan} redraws the two-input additive topology of \texttt{model.plot}, with legible axes, from the archived seed-0 symbolic formulas rather than a visualization refit. Input and output standardizations are inverted. Each edge is centered at its input's training-group mean, with the constant moved into the output intercept; vertical axes therefore show contributions in log-hyperparameter units, not complete reference values. Solid lines mark each input's marginal training range, dashed lines its extrapolation, lower ticks the training-group inputs, and upper triangles OOD inputs. Marginal support does not guarantee joint coverage. The same functions and coordinates serve all three tasks. The original prediction caches are checked for row identity, split separation, finite outputs, and exact reproduction of archived metrics. Source hashes, formulas, the cache audit, and rendering code are retained in \texttt{experiments/kan\_coordinates} and \texttt{figs}.

\subsection{Data efficiency and subset sensitivity}
\label{app:sec2_dataeff}
\begin{figure}[htbp]
\centering
\includegraphics[width=\linewidth]{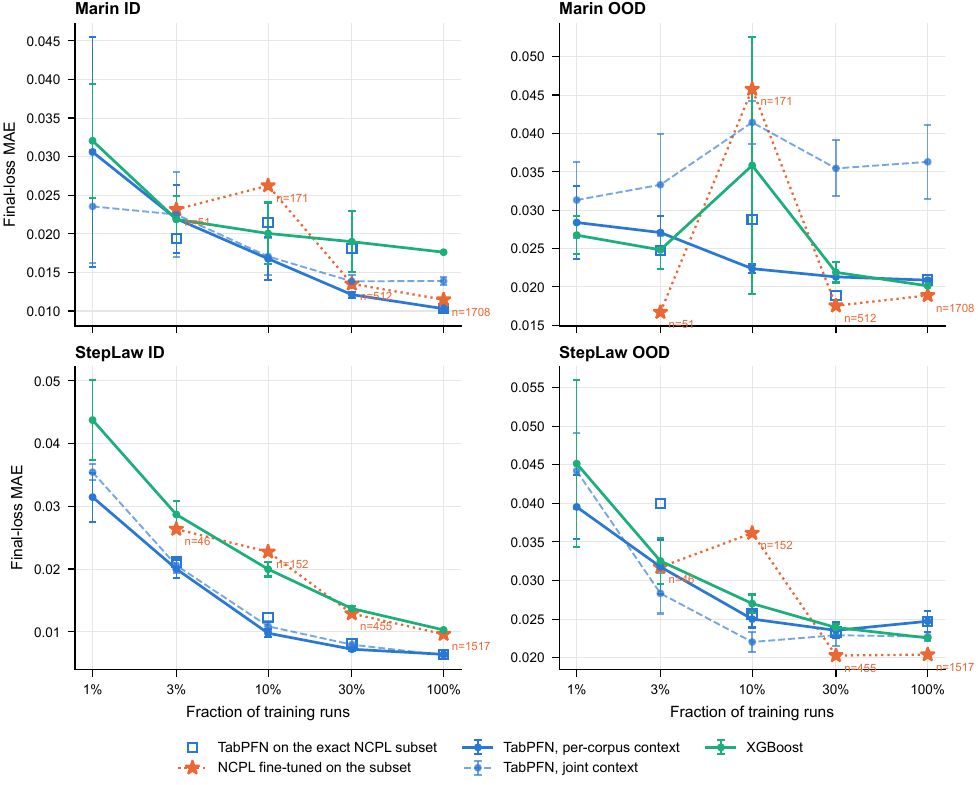}
\caption{\textbf{Data amount and subset composition both affect prediction.} Final-loss MAE versus the fraction of official training runs used as TabPFN context, XGBoost training set, or NCPL fine-tuning set. Stars: NCPL fine-tuned on one sampled subset ($n$ = training runs); squares: per-corpus TabPFN given exactly those runs. Solid and dashed blue lines: TabPFN with per-corpus and joint context on stratified subsets; green: XGBoost. Error bars are standard deviations over three subset draws (over model seeds at 100\%). TabPFN uses the mean readout.}
\label{fig:dataeff}
\end{figure}

Matched-data NCPL fine-tunes use one sampled subset at 3\%, 10\%, and 30\% (released checkpoint at 100\%), and TabPFN receives exactly those records. At 10\% (323 runs), joint-context TabPFN has ID MAE $0.0197/0.0131$ on Marin/StepLaw versus NCPL's $0.0263/0.0227$. With stratified subsets, TabPFN reaches $0.0121$ at 30\% Marin and $0.0098$ at 10\% StepLaw, close to full-data NCPL ($0.0115/0.0096$). Each reduced-data NCPL point is a single fine-tune. Figure~\ref{fig:dataeff} shows the full curves.

\subsection{Input and inference tuning without backbone updates}
\label{app:sec3_inputs}
\label{sec:better_representation}
\label{app:tune}
A greedy ladder tunes encoding, derived features ($\eta\lambda$, warmup fraction, tokens per update, $D/N$, $\eta/B$), feature selection, preprocessing, target transforms, readout, and context routing. The 679-configuration, three-seed search costs 9.9 H200 GPU-hours; ID MAE selects each recipe. The tuned joint recipe lowers Marin final-loss MAE from $0.0093/0.0173$ to $0.0069/0.0146$ (ID/OOD); a recipe tuned on StepLaw alone lowers StepLaw MAE from $0.0063/0.0227$ to $0.0051/0.0190$.

Most of the gain comes from feature construction and selection.

\FloatBarrier

\section{Prospective forecast}
\label{app:forecast_results}
\subsection{Marin Hero Run: frozen forecasts and scoring}
\label{app:hero_protocol}
\paragraph{Chronology.}
V0 was committed on 25 August 2026 at 10:38 UTC, before the next evaluation at 15:43; pre-registration here means a version-controlled timestamp, not an independent registry. We evaluate this first frozen version; later revisions are outside the scope of this paper.

\paragraph{Forecast.}
V0 uses 77 checkpoints from four smaller ladder runs and five hero evaluations through 3.84\% of training. Inputs are log compute, progress, and log active parameters; the target is $\log(L-1.5)$ for dropless Paloma macro loss. TabPFN uses eight estimators, without KAN. Marin's official law fits $L=1.5+AC^{-\alpha}$ to the ladder at each progress value and was archived with V0.

\paragraph{Scoring.}
Outcomes must follow both the forecast timestamp and the last observed step. Off-grid predictions are interpolated piecewise-linearly in $(\log f,\log(L-1.5))$; coverage uses the stored 5th--95th quantiles. On 24 September the public run was marked crashed at step 58,020; the last scored evaluation is step 56,999 (14.61\%, loss $2.2525$). V0 scores 14 outcomes with MAE $0.0216$ and $90\%$ coverage 14/14, versus $0.0576$ for the official law and $0.0788$ for TabPFN given only the ladder. V0's completion forecast ($2.031$, interval $[1.850,2.455]$) remains untested.

\FloatBarrier

\section{Further analyses}
\label{app:further_analyses}
\subsection{Delphi: extrapolating beyond the observed compute range}
\label{app:delphi_meta}
Delphi's 88-run scaling suite \citep{held2026delphi} shares one Qwen3/\allowbreak AdamH/\allowbreak WSD recipe. Its 81 context runs span $3\times10^{18}$--$3\times10^{20}$ FLOPs; the seven held-out runs use $10^{21}$ and $10^{22}$ FLOPs (three seeds each) and $10^{23}$ FLOPs, $333\times$ the largest context budget. Inputs are configuration fields; the target is final Paloma macro loss, residualized by a Chinchilla law fitted on context runs. TabPFN uses eight estimators and three seeds; XGBoost is selected by leave-one-budget-out (LOBO) CV.

\begin{figure}[htbp]
\centering
\includegraphics[width=\linewidth]{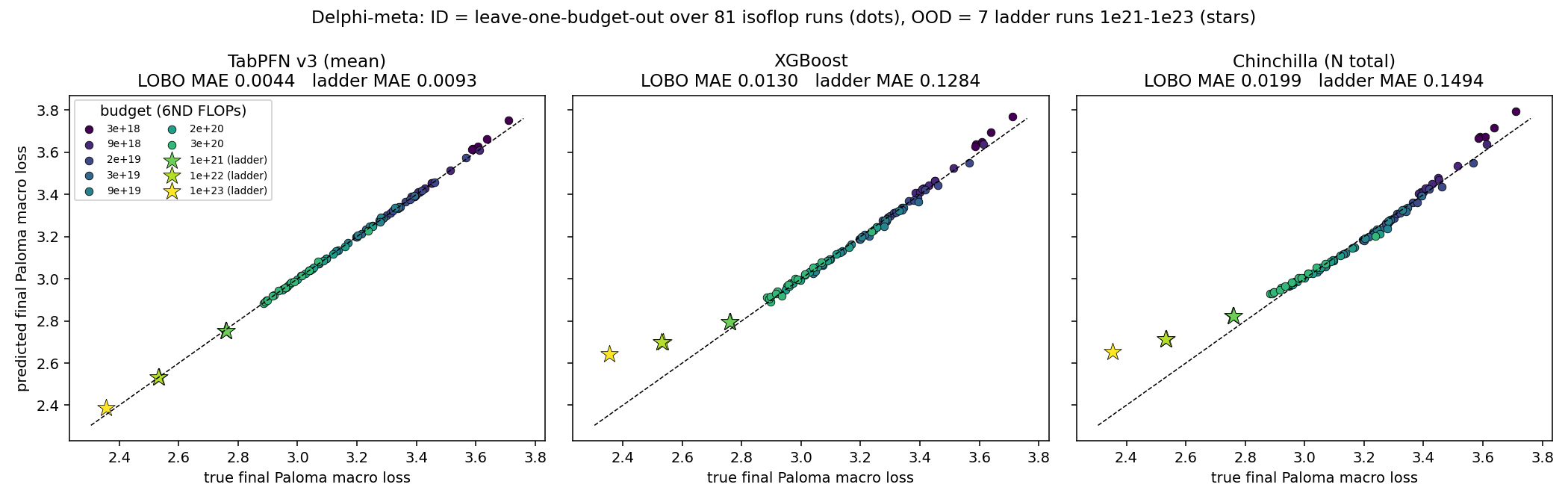}
\caption{\textbf{A held-out scaling ladder.} Observed versus predicted final Paloma macro loss. Dots: leave-one-budget-out predictions in the context pool; stars: held-out ladder runs (residual-mean TabPFN).}
\label{fig:delphi_meta}
\end{figure}

On the seven-run ladder, residual-mean TabPFN has MAE $0.0093$ nats ($0.005/0.006/0.032$ at $10^{21}/10^{22}/10^{23}$ FLOPs), versus $0.0182$ for the Delphi team's prior forecast, $0.128$ for XGBoost, $0.149$ for the Chinchilla fit, and $0.204$ for released NCPL (Figure~\ref{fig:delphi_meta}). Within the 81-run pool, leave-one-budget-out and leave-one-width-out errors are $0.0044$ and $0.0040$; without residualization, ladder error rises to $0.053$.

\FloatBarrier
\subsection{Changing feature order without changing a training configuration}
\label{app:permutation}
We apply eight fixed permutations to TabPFN's columns (context and query alike) and to NCPL's serialized key--value pairs, on the 1,905 Marin/StepLaw ID and OOD runs. TabPFN uses one estimator and three seeds, separating external permutations from its ensemble.

Across the four splits, the permutation-averaged MAE of NCPL is $1.8$--$4.0\times$ its canonical MAE (for example, $0.0115\to0.0337$ on Marin ID), rising under every permutation, whereas TabPFN stays within $0.97$--$1.08\times$.

\FloatBarrier
\subsection{Row embeddings of the joint-context model}
\label{app:embed}
\begin{figure}[htbp]
\centering
\includegraphics[width=\linewidth]{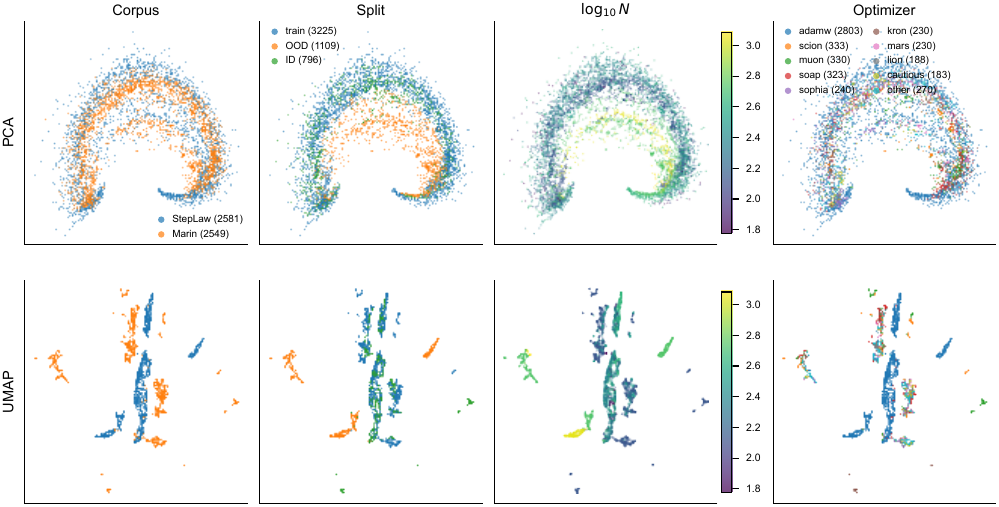}\\[4pt]
\includegraphics[width=\linewidth]{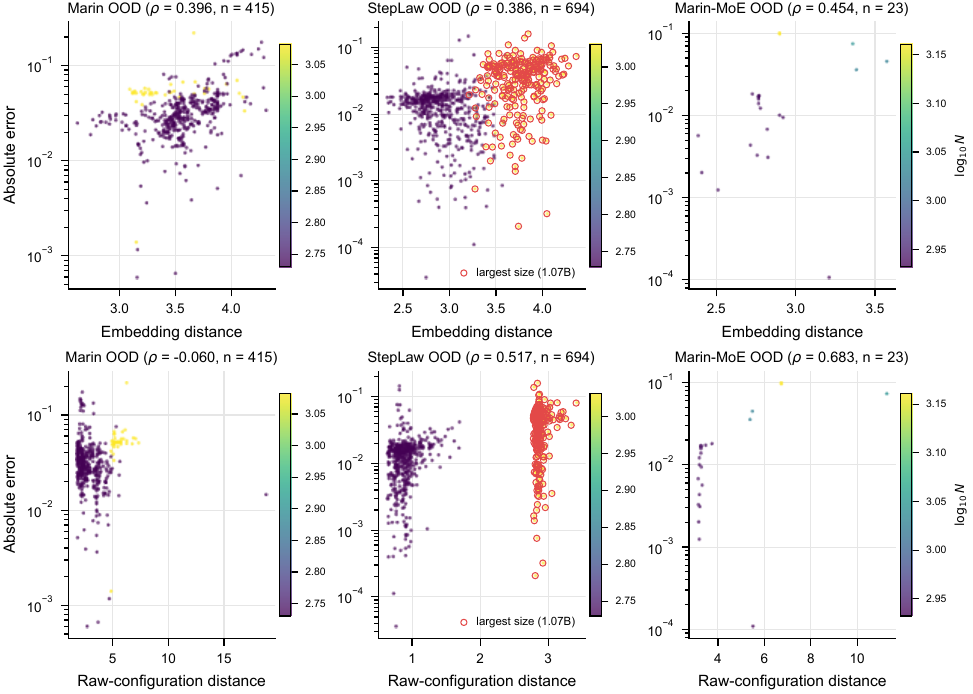}
\caption{\textbf{Row embeddings of the joint-context model.} Top: PCA (first row) and UMAP \citep{mcinnes2018umap} (second row) of all 5,130 runs, colored by corpus, split, $\log_{10}N$, and optimizer. Bottom: distance to the ten nearest context rows versus absolute error for OOD runs, in embedding space (first row) and raw configuration space (second row); red rings mark the largest StepLaw size (1.07B).}
\label{fig:embed}
\end{figure}

\begin{figure}[htbp]
\centering
\includegraphics[width=\linewidth]{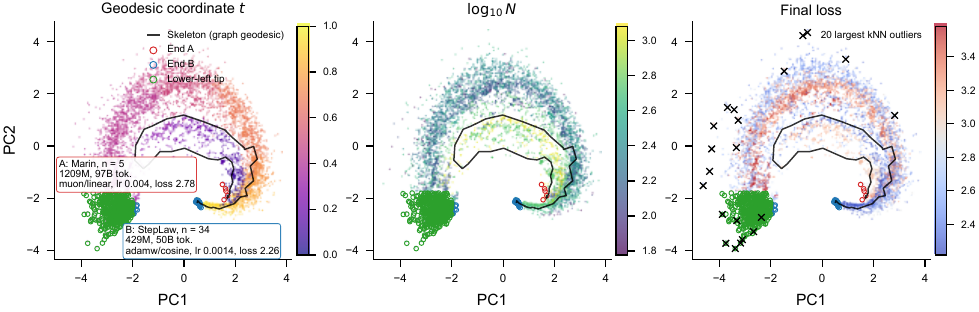}
\caption{\textbf{Probing the horseshoe.} The same PCA colored by the geodesic coordinate $t$ (left; the skeleton and its endpoints are annotated with their median configurations), by $\log_{10}N$ (middle; size varies radially, not along the arc), and by final loss (right; crosses mark the 20 largest $k$-nearest-neighbor outliers).}
\label{fig:manifoldcorners}
\end{figure}

TabPFN also yields a context-conditioned 512-dimensional embedding for each run (out-of-fold for context rows). Figure~\ref{fig:embed} (top) projects all 5,130 Marin and StepLaw runs of the joint context. The runs form a horseshoe: the two corpora share the arc but remain locally separated (10-nearest-neighbor same-source purity $0.985$), and model size varies radially, with the larger held-out OOD runs on the inner band. Final loss varies along the arc, reaching its lowest values at the lower-right hook of compute-rich StepLaw runs (429M parameters, 50B tokens; Figure~\ref{fig:manifoldcorners}). Along a graph geodesic through the arc, deviation from the fitted scaling law decreases (Spearman $-0.50$), whereas learning rate, batch size, and weight decay are nearly uncorrelated with arc position ($|\rho|\le0.05$).

Distance to the context in this embedding tracks OOD error on all three corpora (Figure~\ref{fig:embed}, bottom; Spearman $0.40$, $0.39$, and $0.45$ on Marin, StepLaw, and MoE). The largest StepLaw models are both the farthest from the context and the worst predicted.

\FloatBarrier
\subsection{Training responses: recovering known relationships}
\label{sec:laws}
We vary one configuration field at a time and compare each predictor's response with measured sweeps (about 5,600 queries, seed 0, per-corpus mean predictions). These probe a surrogate, not new training runs.

TabPFN and NCPL both reproduce learning-rate U-curves at all 31 anchors, the optimal-learning-rate exponent ($-0.64$; measured $-0.643$), and batch-size saturation beyond 256; the predicted batch-optimum exponent in token budget is $0.55$ versus $0.62$ measured.

\FloatBarrier
\subsection{A proposed canonical training record}
\label{app:training_record_schema}
Main experiments use corpus-specific flat encodings. A proposed canonical schema (133 columns: 67 values and 66 masks) distinguishes recorded, inapplicable, and missing fields and fixes per-corpus units.

\end{document}